%% file: main.tex
\pdfoutput=1
\documentclass[10pt,twocolumn,letterpaper]{article}

\usepackage{iccv}               

\usepackage{graphicx}  
\usepackage{amsmath}   
\usepackage{amssymb}   
\usepackage{epsfig}
\usepackage{bm}        
\usepackage{bbm}       
\usepackage{mathtools} 
\usepackage{xcolor}    
\usepackage{soul}      
\usepackage{booktabs}  
\usepackage{enumitem}  
\usepackage{tabularx}  
\usepackage{multirow}

\input{defs.tex}

\usepackage[pagebackref=false,breaklinks=false,%
            colorlinks=true,bookmarks=true,citecolor=ourdarkblue,%
            urlcolor=ourdarkblue,linkcolor=ourdarkblue]{hyperref}

\def\iccvPaperID{4700} 

\begin{document}

\title{Synthetic Data from Diffusion Models\\ Improves ImageNet Classification}

\author{Shekoofeh Azizi,
Simon Kornblith, 
Chitwan Saharia\textsuperscript{*},
Mohammad Norouzi\textsuperscript{\thanks{Work done at Google Research.}},
David J.\ Fleet\\[.12cm]
Google Research, Brain Team \thanks{{\{shekazizi,\,skornblith,\,davidfleet\}@google.com}}\:
}

\maketitle

\begin{abstract}
\vspace*{-0.35cm}
Deep generative models are becoming increasingly powerful, now generating diverse high fidelity photo-realistic samples given text prompts. Have they reached the point where models of natural images can be used for generative data augmentation, helping to improve challenging discriminative tasks? 
We show that large-scale text-to-image diffusion models can be fine-tuned to produce class-conditional models with SOTA FID (1.76 at $256\!\times\! 256$ resolution) and Inception Score 
(239 at $256\times 256$).
The model also yields a new SOTA in Classification Accuracy Scores (64.96 for $256\!\times\! 256$ generative samples, improving to 69.24 for $1024\!\times\! 1024$ samples).
Augmenting the ImageNet training set with samples from the resulting models yields significant improvements in ImageNet classification accuracy over strong ResNet and Vision Transformer baselines.

\vspace*{-0.15cm}
\end{abstract}

\input{introduction}

\input{background}

\input{experiments}
\input{results}


\input{conclusion}


{\small
\bibliographystyle{ieee_fullname}
\bibliography{references}
}

\newpage

\input{appendix}

\end{document}

%% file: defs.tex
\definecolor{ourlightblue}{HTML}{C7EBFF}         
\definecolor{ourlightorange}{HTML}{FFDBC3}       
\definecolor{ourlightgray}{HTML}{EAEAF2}         
\definecolor{ourlightgreen}{HTML}{BDF0E2}        
\definecolor{ourdarkblue}{HTML}{0033CC}          
\definecolor{ourdarkgreen}{HTML}{28B23D}         
\definecolor{ourdarkred}{HTML}{ED1C1C}            %


%% file: introduction.tex
\vspace*{-0.05cm}
\section{Introduction}
\vspace*{-0.1cm}

Deep generative models are becoming increasingly mature to the point that they can generate high fidelity photo-realistic samples \cite{dhariwal2021diffusion,ho2020denoising,sohl2015deep}. 
Most recently, denoising diffusion probabilistic models (DDPMs) \cite{ho2020denoising,sohl2015deep} have emerged as a new category of generative techniques that are capable of generating images comparable to generative adversarial networks (GANs) in quality while introducing greater stability during training
\cite{dhariwal2021diffusion,ho2022cascaded}. 
This has been shown both for class-conditional generative models on classification datasets, and for open vocabulary text-to-image generation \cite{nichol2021glide,ramesh2022hierarchical,rombach2022high,saharia2022photorealistic}.

It is therefore natural to ask whether current models are powerful enough to generate natural image data that are effective for challenging discriminative tasks; i.e., {\em  generative data augmentation}.
Specifically, are diffusion models capable of producing image samples of sufficient quality and diversity to improve performance on well-studied benchmark tasks like ImageNet classification?
Such tasks set a high bar, since existing architectures, augmentation strategies, and training recipes have been heavily tuned.
A closely related question is, to what extent large-scale text-to-image models can serve as good representation learners or foundation models for downstream tasks? We explore this issue in the context of generative data augmentation, showing that these models can be fine-tuned to produce state-of-the-art class-conditional generative models on ImageNet.

\begin{figure}[t]
\begin{center}
\includegraphics[width=0.43\textwidth]{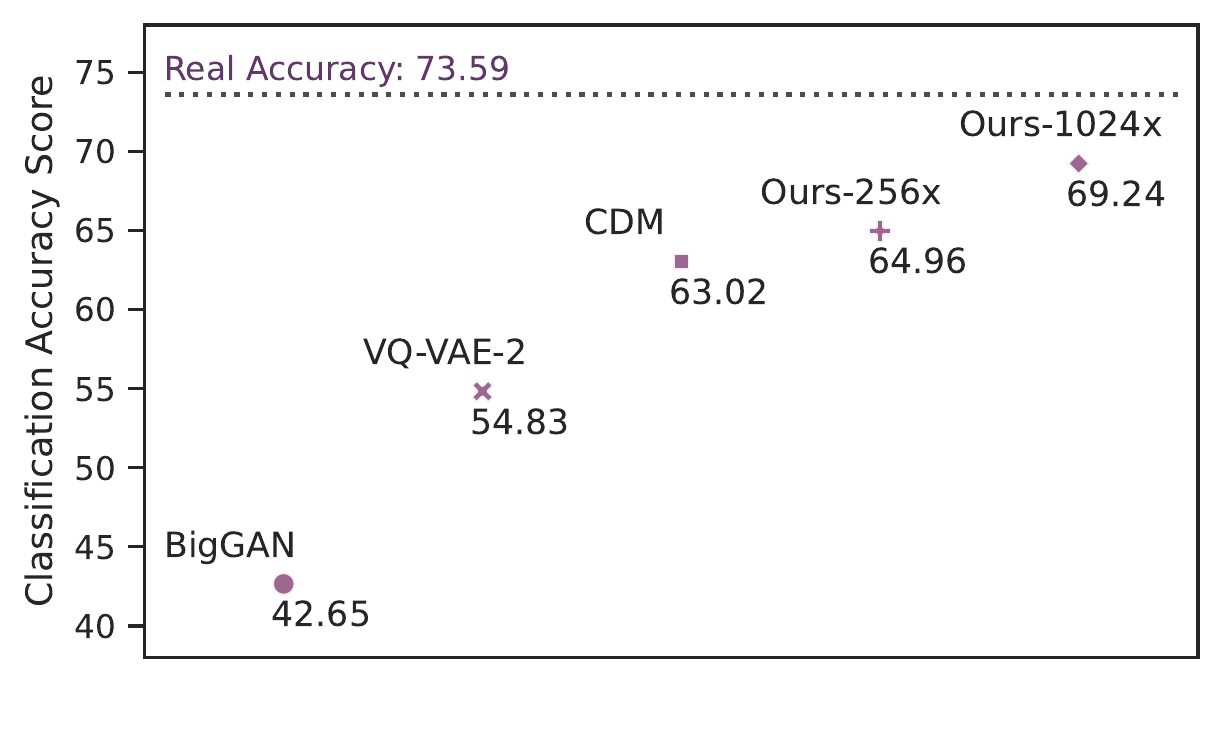}\\
\vspace*{-0.35cm}
\includegraphics[width=0.42\textwidth]{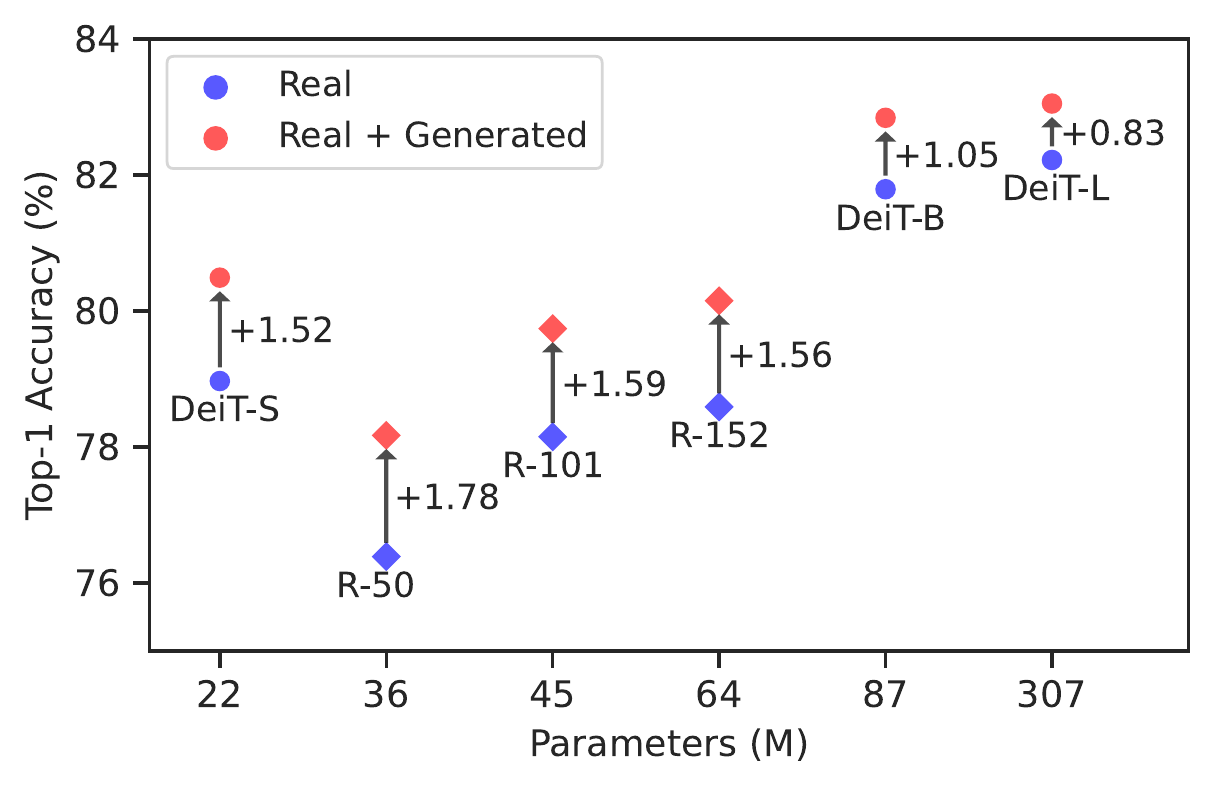}
\end{center}
\vspace*{-0.45cm}
\caption{Top: Classification Accuracy Scores \cite{ravuri2019classification} show that models trained on generated data are approaching those trained on real data. 
Bottom: Augmenting real training data with generated images from our ImageNet model boosts classification accuracy for  ResNet and  Transformer models.}
\vspace*{-0.4cm}
\label{fig:teaser}
\end{figure}

\begin{figure*}[t]
\begin{center}
\includegraphics[width=0.89\textwidth]{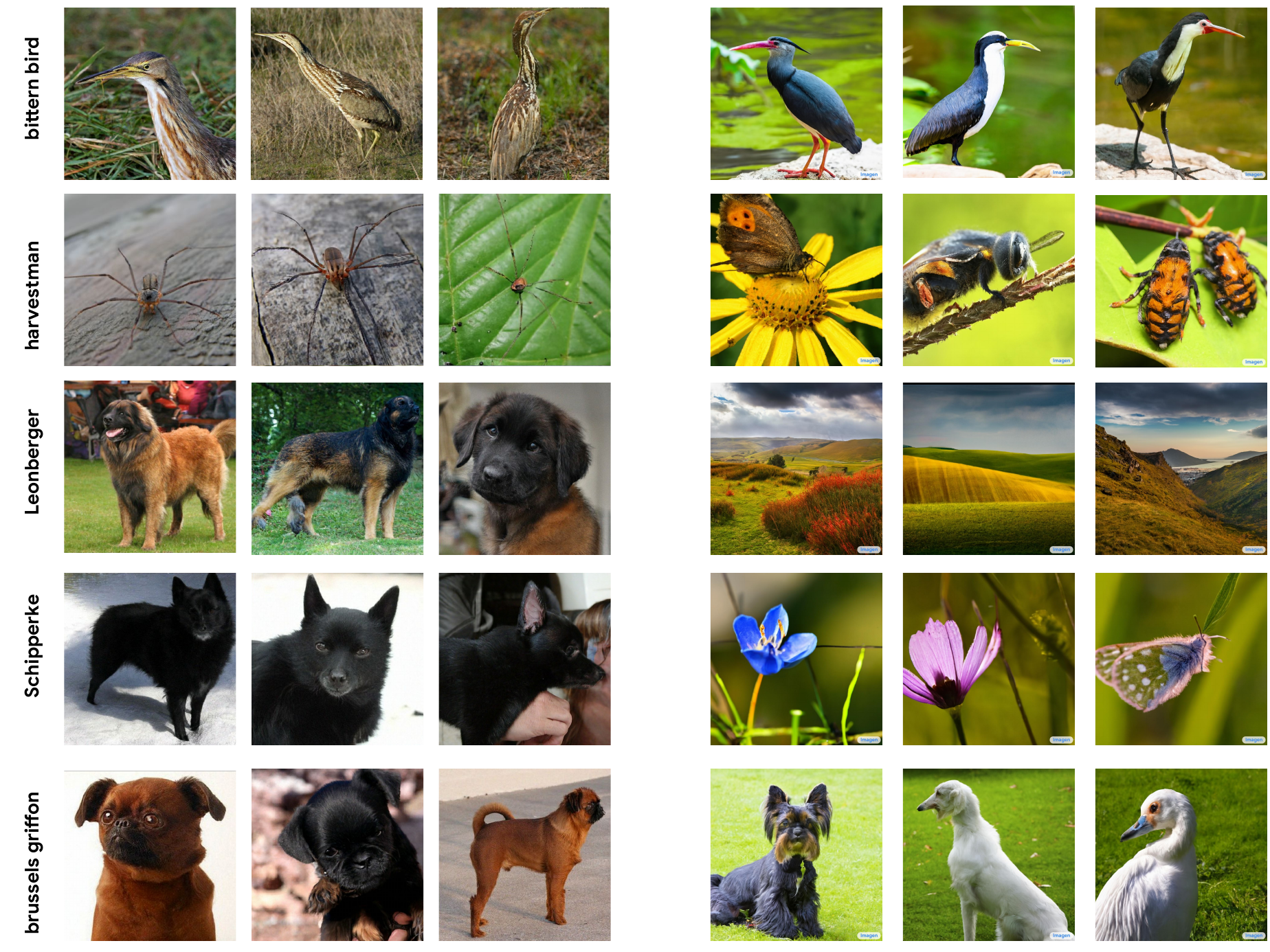}
\end{center}
\vspace*{-0.4cm}
\caption{Example $1024\!\times\! 1024$ images from the fine-tuned Imagen (left) model vs.\ vanilla Imagen (right). Fine-tuning and careful choice of guidance weights and other sampling parameters help to improve the alignment of images with class labels and sample diversity. 
More samples are provide in the Appendix.
}
\vspace*{-0.4cm}
\label{fig:samples}
\end{figure*}

To this end, we demonstrate three key findings. First, we show that an Imagen model fine-tuned on ImageNet training data produces state-of-the-art class-conditional ImageNet models at multiple resolutions, according to their Fréchet Inception Distance (FID)~\cite{heusel2017gans} and Inception Score (IS)~\cite{salimans2016improved}; e.g,, we obtain an FID of 1.76 and IS of 239 on $256\!\times\! 256$ image samples.
These models outperform existing state-of-the-art models, with or without the use of guidance to improve model sampling.
We further establish that data from such fine-tuned class-conditional models also provide new state-of-the-art Classification Accuracy Scores (CAS) \cite{ravuri2019classification}, computed by training ResNet-50 models on synthetic data and then evaluating them on the real ImageNet validation set (Fig. \ref{fig:teaser} - top).
Finally, we show that performance of models trained on generative data further improves by combining synthetic data with real data, with larger amounts of synthetic data, and with longer training times.  These results hold across a host of convolutional and Transformer-based architectures (Fig.~\ref{fig:teaser} - bottom).

%% file: background.tex
\vspace*{-0.1cm}
\section{Related Work}
\vspace*{-0.05cm}

\noindent
\textbf{Synthetic Data.} The use of synthetic data has been widely explored for generating large amounts of labeled data for vision tasks that require extensive annotation.
Examples include tasks like semantic image segmentation 
 \cite{baranchuk2021label,chen2019learning,li2022bigdatasetgan,li2021semantic,sankaranarayanan2018learning, tritrong2021repurposing}, optical flow estimation \cite{dosovitskiy2015flownet,kim2022transferable,sun2021autoflow}, human motion understanding \cite{guo2022learning,izadi2011kinectfusion,ma2022pretrained,varol2017learning}, and other dense prediction tasks \cite{baranchuk2021label,xu2021generative}. Previous work has explored 3D-rendered datasets \cite{greff2021kubric,zheng2020structured3d} and simulation environments with physically realistic engines \cite{de2021next,dosovitskiy2017carla,gan2021threedworld}. 
Unlike methods that use model-based rendering, here we focus on the use of data-driven generative models of natural images, for which GANs have remained the predominant approach to date \cite{brock2019large,gowal2021improving,li2022bigdatasetgan}. Very recent work has also explored the use of publicly available text-to-image diffusion models to generate synthetic data. We discuss this work further below.

\noindent
\textbf{Distillation and Transfer.} In our work, we use a diffusion model that has been pretrained on a large multimodal dataset and fine-tuned on ImageNet to provide synthetic data for a classification model. This setup has connections to previous work that has directly trained classification models on large-scale datasets and then fine-tuned them on ImageNet~\cite{kolesnikov2020big,mahajan2018exploring,radford2021learning,sun2017revisiting,zhai2022scaling}. 
It is also related to knowledge distillation~\cite{bucilua2006model,hinton2015distilling} in that we transfer knowledge from the diffusion model to the classifier, although it differs from the traditional distillation setup in that we transfer this knowledge through generated data rather than labels. Our goal in this work is to show the viability of this kind of generative knowledge transfer with modern diffusion models.

\noindent
\textbf{Diffusion Model Applications.} 
Diffusion models have been successfully applied to image generation \cite{ho2020denoising,ho2022cascaded,ho2022classifier}, speech generation \cite{chen2020wavegrad,kong2020diffwave}, and video generation \cite{ho2022imagen,singer2022make,villegas2022phenaki}, and have found applications in various image processing areas, including image colorization, super-resolution, inpainting, and semantic editing \cite{saharia2022palette,saharia2022image,song2020score,wang2022imagen}. One notable application of diffusion models is large-scale text-to-image generation. Several text-to-image models including Stable Diffusion~\cite{rombach2022high}, DALL-E 2~\cite{ramesh2022hierarchical}, Imagen~\cite{saharia2022photorealistic}, eDiff \cite{eDiff}, and GLIDE~\cite{nichol2021glide} have produced evocative high-resolution images. 
However, the use of large-scale diffusion models to support downstream tasks is still in its infancy.

Very recently, large-scale text-to-image models have been used to augment training data. 
He et al.~\cite{he2022synthetic} show that synthetic data generated with GLIDE~\cite{nichol2021glide} improves zero-shot and few-shot image classification performance. They further demonstrate that a synthetic dataset generated by fine-tuning GLIDE on CIFAR-100 images can provide a substantial boost to CIFAR-100 classification accuracy. Trabucco et al.~\cite{trabucco2023effective} explore strategies to augment individual images using a pretrained diffusion model, demonstrating improvements in few-shot settings. Most closely related to our work, two recent papers train ImageNet classifiers on images generated by diffusion models, although they explore only the pretrained Stable Diffusion model and do not fine-tune it~\cite{bansal2023leaving,sariyildiz2022fake}. They find that images generated in this fashion do not improve accuracy on the clean ImageNet validation set. Here, we show that the Imagen text-to-image model can be fine-tuned for class-conditional ImageNet, yielding SOTA models.

\vspace*{-0.05cm}
\section{Background}
\label{sec:background}

\noindent
\textbf{Diffusion.} 
Diffusion models work by gradually destroying the data through the successive addition of Gaussian noise, and then learning to recover the data by reversing this noising process~\cite{ho2020denoising,sohl2015deep}. 
In broad terms, in a forward process random noise is slowly added to the data as time $t$ increases from 0 to $T$. 
A learned reverse process inverts the forward process, gradually refining a sample of noise into an image.
To this end, samples at the current time step, $x_{t-1}$ are drawn from a learned Gaussian distribution $\mathcal{N}( x_{t-1}; \mu_\theta (x_t, t), \Sigma_\theta(x_t, t))$ where the mean of the distribution $\mu_\theta (x_t, t)$, is conditioned on the sample at the previous time step. The variance of the distribution $\Sigma_\theta(x_t, t)$ follows a fixed schedule. In conditional diffusion models, the reverse process is associated with a conditioning signal, such as a class label in class-conditional models \cite{ho2022cascaded}.

Diffusion models have been the subject of many recent papers including important innovations in architectures and training (e.g.,~\cite{eDiff,ho2022cascaded,nichol2021improved,saharia2022photorealistic}). 
Important below, \cite{ho2022cascaded} propose cascades of diffusion models at increasing image resolutions for high resolution images.
Other work has explored the importance of the generative sampling process, introducing new noise schedules, guidance mechanisms to trade-off diversity with image quality, distillation for efficiency, and different parameterizations of the denoising objective (e.g., \cite{hoogeboom2023simple,KarrasNeurIPS2022,saharia2022photorealistic, salimans2022progressive}).

\vspace*{0.05cm}
\noindent
\textbf{Classification Accuracy Score.} It is a standard practice to use FID~\cite{heusel2017gans} and Inception Score~\cite{salimans2016improved} to evaluate the visual quality of generative models. 
Due to their relatively low computation cost, these metrics are widely used as proxies for generative model training and tuning. 
However, both methods tend to penalize non-GAN models harshly, and Inception Score produces overly optimistic scores in methods with sampling modifications~\cite{ho2022classifier,ravuri2019classification}. 
More importantly, Ravuri and Vinyals~\cite{ravuri2019classification} argued that these metrics do not show a consistent correlation with metrics that assess performance on downstream tasks like classification accuracy.

An alternative way to evaluate the quality of samples from generative models is to examine the performance of a classifier that is trained on generated data and evaluated on real data~\cite{santurkar2018classification,yang2017lrgan}. To this end, Ravuri and Vinyals~\cite{ravuri2019classification} propose classification accuracy score (CAS), which measures classification performance on the ImageNet validation set for ResNet-50 models \cite{he2016deep} trained on generated data. It is an intriguing proxy, as it requires generative models to produce high fidelity images across a broad range of categories, competing directly with models trained on real data. 

To date, CAS performance has not been particularly compelling.
Models trained exclusively on generated samples underperform those trained on real data. Moreover, performance drops when even relatively small amounts of synthetic data are added to real data during training \cite{ravuri2019classification}.
This performance drop may arise from issues with the quality of generated sample, the diversity (e.g., due to mode collapse in GAN models), or both.
Cascaded diffusion models \cite{ho2022cascaded} have recently been shown to outperform BigGAN-deep \cite{brock2019large}
VQ-VAE-2 \cite{razavi2019generating} on CAS (and other metrics).
That said, there remains a sizeable gap in ImageNet test performance between models trained on real data and those trained on synthetic data \cite{ho2022cascaded}.
Here, we explore the use of diffusion models in greater depth, with much stronger results,  demonstrating the advantage of large-scale models and fine-tuning.

%% file: experiments.tex
\begin{figure*}[t]
\begin{center}
\includegraphics[width=0.33\textwidth]{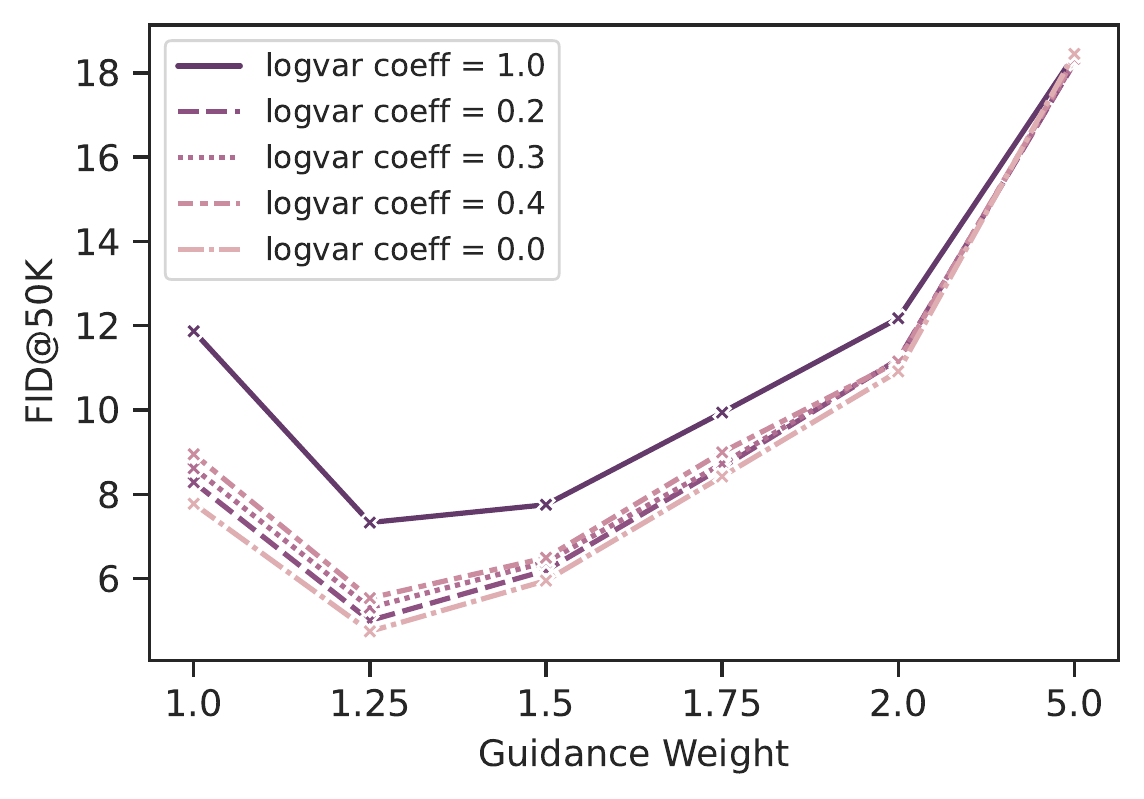}
\includegraphics[width=0.33\textwidth]{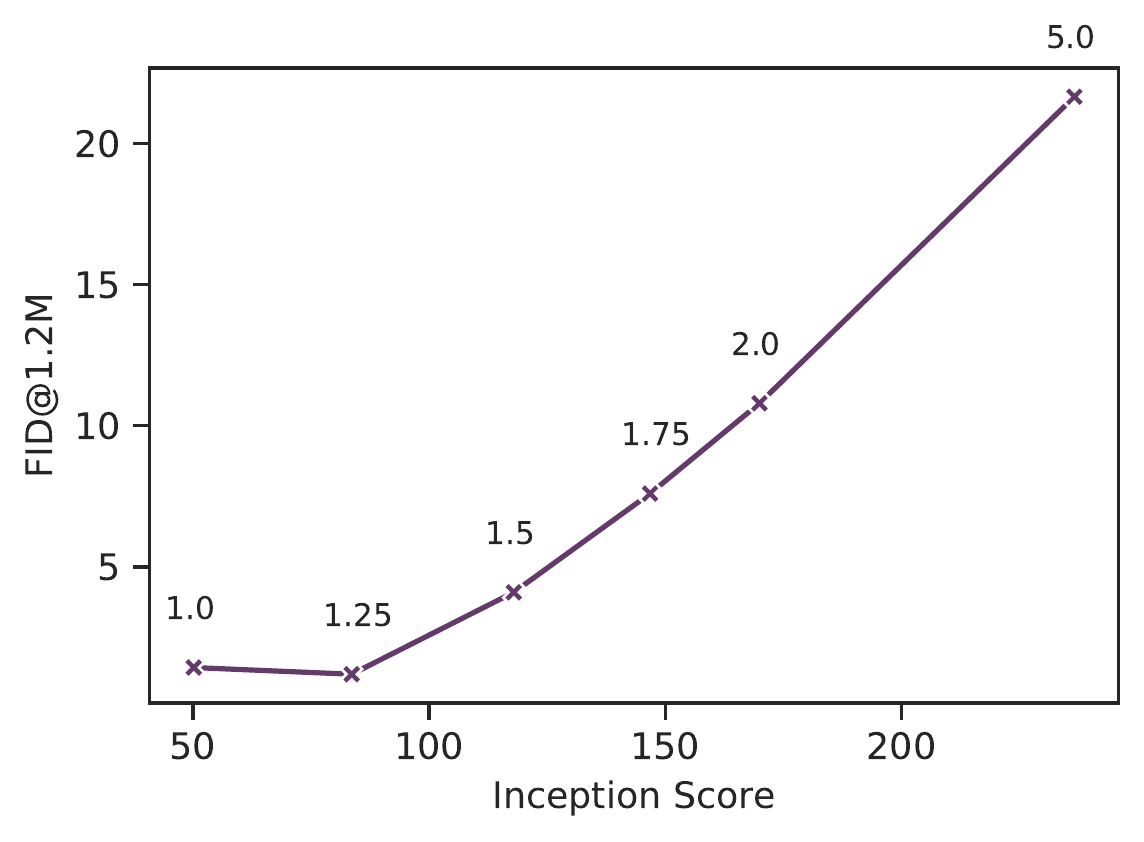}
\includegraphics[width=0.33\textwidth]{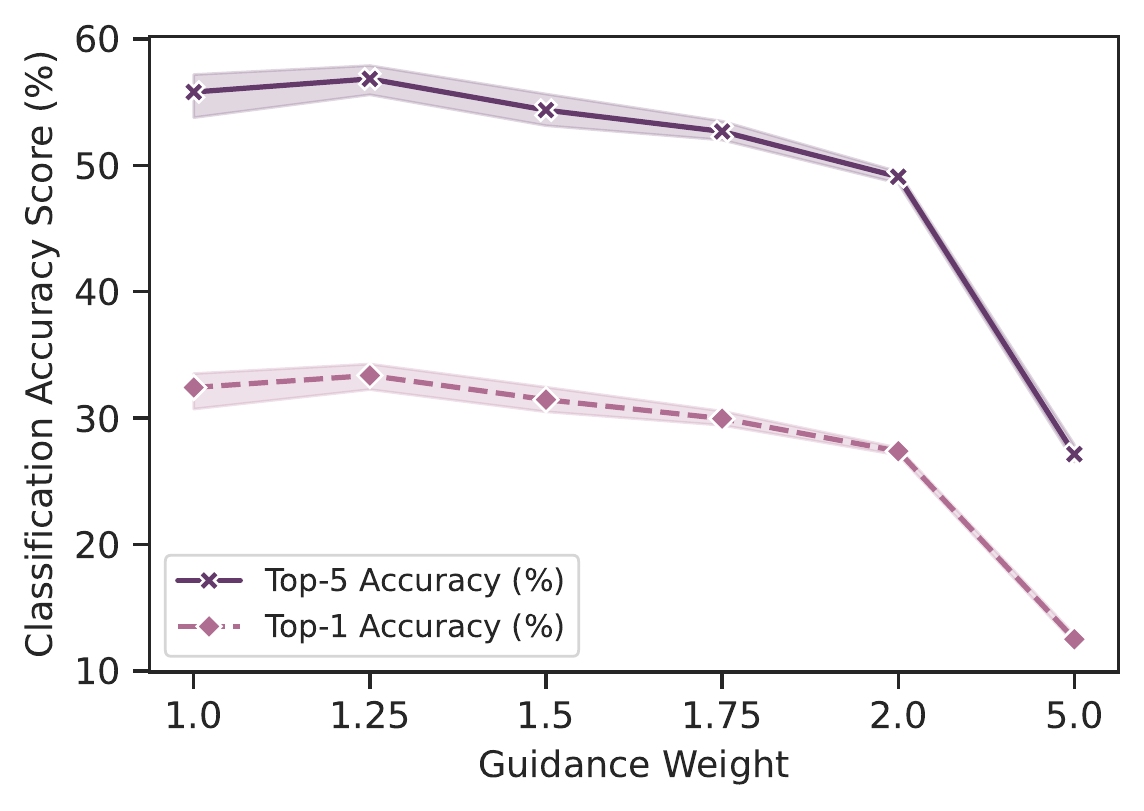}
\end{center}
\vspace*{-0.4cm}
\caption{Sampling refinement for $64\!\times\!64$ base model. \textbf{Left}: Validation set FID vs.\ guidance weights for different values of log-variance. \textbf{Center}: Pareto frontiers for training set FID and IS at different values of the guidance weight. \textbf{Right}: Dependence of CAS on guidance weight.} 
\label{fig:fid_is_cas_pareto_64x}
\vspace*{-0.4cm}
\end{figure*}

\vspace*{-0.05cm}
\section{Generative Model Training and Sampling}
\vspace*{-0.1cm}

In what follows we address two main questions: whether large-scale text-to-image models can be fine-tuned as class-conditional ImageNet models, and to what extent such models are useful for generative data augmentation. 
For this purpose, we undertake a series of experiments to construct and evaluate such models, focused primarily on data sampling for use in training ImageNet classifiers.
ImageNet classification accuracy is a high bar as a domain for generative data augmentation as the task is widely studied, and existing architectures and training recipes are very well-honed.

The ImageNet ILSVRC 2012 dataset~\cite{russakovsky2015imagenet} (ImageNet-1K) comprises 1.28 million labeled training images and 50K validation images spanning 1000 categories. We adopt ImageNet-1K as our benchmark to assess the efficacy of the generated data, as this is one of the most widely and thoroughly studied benchmarks for which there is an extensive literature on architectures and training procedures, making it challenging to improve performance. 
Since the images of ImageNet-1K dataset vary in dimensions and resolution with the average image resolution of 469$\times$387~\cite{russakovsky2015imagenet}, we examine synthetic data generation at different resolutions, including 64$\times$64, 256$\times$256, and 1024$\times$1024. 

In contrast to previous work that trains diffusion models directly on ImageNet data (e.g., \cite{dhariwal2021diffusion,ho2022cascaded,hoogeboom2023simple}), here we leverage a large-scale text-to-image diffusion model \cite{saharia2022photorealistic} as a foundation, in part to explore the potential benefits of pre-training on a larger, generic corpus.
A key challenge in doing so concerns the alignment of the text-to-image model with ImageNet classes.
If, for example, one naively uses short text descriptors like those produced for CLIP by \cite{radford2021learning} as text prompts for each ImageNet class, 
the data generated by the Imagen models is easily shown to produce very poor ImageNet classifier. One problem is that a given text label may be associated with multiple visual concepts in the wild, or visual concepts that differ systematically from ImageNet (see Figure \ref{fig:samples}).  This poor performance may also be a consequence of the high guidance weights used by Imagen, effectively sacrificing generative diversity for sample quality.
While there are several ways in which one might re-purpose a text-to-image model
as a class-conditional model, e.g., optimizing prompts in order to minimize
the distribution shift, here we fix the prompts
to be the one or two words class names from \cite{radford2021learning}, and 
fine-tune the weights and sampling parameters of the diffusion-based generative model.

\subsection{Imagen Fine-tuning}
\vspace*{-0.1cm}

We leverage the large-scale Imagen text-to-image model described in detail in \cite{saharia2022photorealistic} as the backbone text-to-image generator that we fine-tune using the ImageNet training set. 
It includes a pretrained text encoder that maps text to contextualized embeddings, and a cascade of conditional diffusion models that map these embeddings to images of increasing resolution. 
Imagen uses a frozen T5-XXL encoder as a semantic text encoder to capture the complexity and compositionality of text inputs. 
The cascade begins with a 2B parameter 64$\times$64 text-to-image base model.
Its outputs are then fed to a 600M parameter super-resolution model to upsample from 64$\times$64 to 256$\times$256, followed by a 400M parameter model to upsample from 256$\times$256 to 1024$\times$1024. 
The base 64$\times$64 model is conditioned on text embeddings via a pooled embedding vector added to the diffusion time-step embedding, like previous class-conditional diffusion models \cite{ho2022cascaded}. All three stages of the diffusion cascade include text cross-attention layers \cite{saharia2022photorealistic}. 

Given the relative paucity of high resolution images in ImageNet, we fine-tune only the 64$\times$64 base model and 64$\times$64$\rightarrow$256$\times$256 super-resolution model
on the ImageNet-1K train split, keeping the final super-resolution module and text-encoder unchanged. 
The 64$\times$64 base model is fine-tuned for 210K steps and the  64$\times$64$\rightarrow$256$\times$256 super-resolution model is fine-tuned for 490K steps, on 256 TPU-v4 chips with a batch size of 2048. 
As suggested in the original Imagen training process, Adafactor \cite{adafactor} is used to fine-tune the base 64$\times$64 model because it had a smaller memory footprint compared to Adam \cite{adam}. 
For the 256$\times$256 super-resolution model, we used Adam for better sample quality.
Throughout fine-tuning experiments, we select models based on FID score calculated over 10K samples from the default Imagen sampler and the ImageNet-1K validation set.

\subsection{Sampling Parameters}
\vspace*{-0.1cm}

The quality, diversity, and speed of text-conditioned diffusion model sampling are strongly affected by multiple factors including the number of diffusion steps, noise condition augmentation \cite{saharia2022photorealistic}, 
guidance weights for classifier-free guidance \cite{ho2022classifier,nichol2021glide},
and the log-variance mixing coefficient used for prediction (Eq.\ 15 in \cite{nichol2021improved}), described in further detail in Appendix~\ref{app:Imagen-HPs}.
We conduct a thorough analysis of the dependence of FID, IS and classification accuracy scores (CAS) in order to select good sampling parameters for the downstream classification task.

\begin{figure}
\begin{center}
\includegraphics[width=0.33\textwidth]{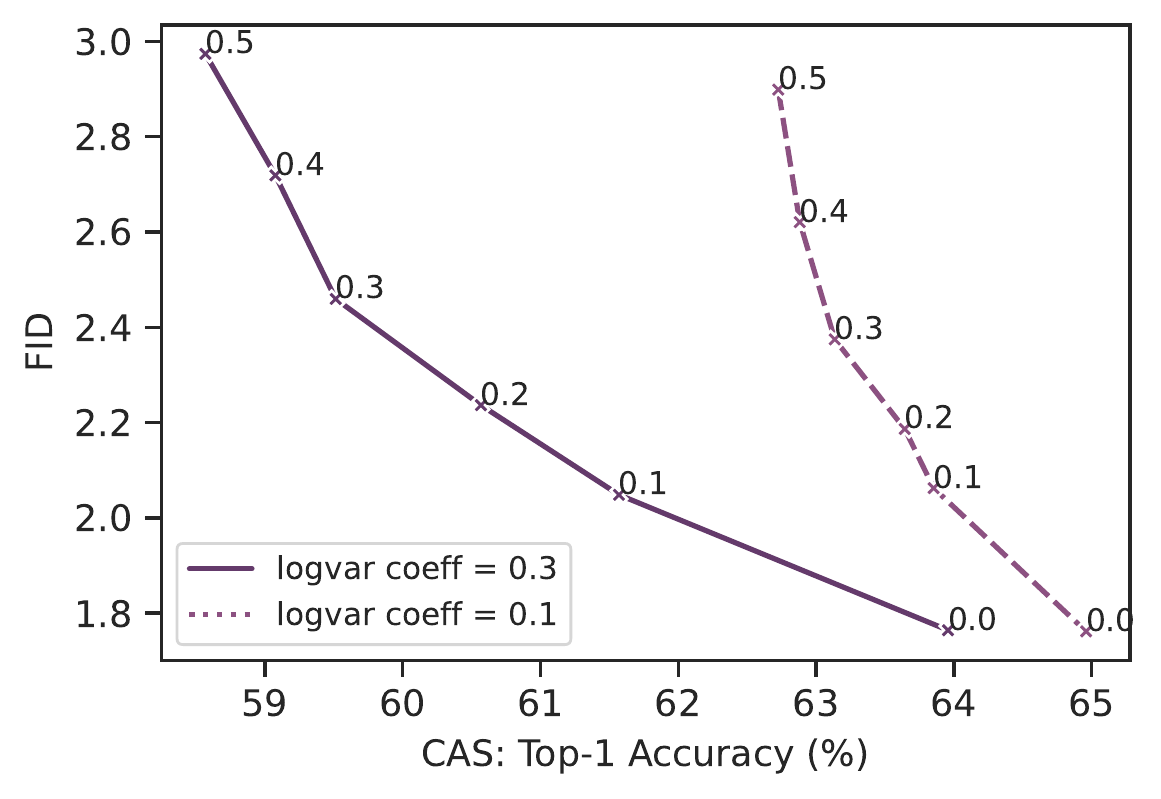}
\end{center}
\vspace*{-0.35cm}
\caption{Training set FID vs.\ CAS Pareto curves under varying noise conditions when the guidance weight is set to 1.0 for resolution $256\!\times\!256$. These curves depict the joint influence of the log-variance mixing coefficient \cite{nichol2021improved} and noise conditioning augmentation \cite{ho2022cascaded} on  FID and CAS.}
\label{fig:fid_cas_log_var_256x}
\vspace*{-0.2cm}
\end{figure}

The sampling parameters for the $64\times 64$ based model establish the overall quality and diversity of image samples. 
We first sweep over guidance weight, log-variance, and number of sampling steps, to identify good hyperparameters based on FID-50K (vs.\ the ImageNet validation set).
Using the DDPM sampler \cite{ho2020denoising} for the base model, we sweep over guidance values of $\left [ 1.0, 1.25, 1.5, 1.75, 2.0, 5.0 \right ]$ and log-variance of $\left [ 0.0, 0.2, 0.3, 0.4, 1.0 \right ]$, and denoise for 128, 500, or 1000 steps.
The results of this sweep, summarized in Figure~\ref{fig:fid_is_cas_pareto_64x}, suggest that optimal FID is obtained with a log-variance of 0 and 1000 denoising steps.
Given these parameter choices we then complete a more compute intensive sweep, sampling 1.2M images from the fine-tuned base model for different values of the guidance weights. We measure FID, IS and CAS for these samples on the validation set in order
to select the guidance weight for the model.
Figure~\ref{fig:fid_is_cas_pareto_64x} shows the Pareto frontiers for FID vs.\ IS across different guidance weights, as well as 
the dependence of CAS on guidance weight, suggesting that optimal FID and CAS are obtained at a guidance weight of 1.25.

Given $64\times 64$ samples obtained with the optimal hyperparameters, we then analyze the impact of guidance weight, noise augmentation, and log-variance to select sampling parameters for the super-resolution models. 
The noise augmentation value specifies the level of noise augmentation applied to the input to super-resolution stages  in the Imagen cascade to regulate sample diversity (and improve robustness during model training).  
Here, we sweep over guidance values of $\left [ 1.0, 2.0, 5.0, 10.0, 30.0 \right ]$, noise conditioning augmentation values of $\left [ 0.0, 0.1, 0.2, 0.3, 0.4 \right ]$, and log-variance mixing coefficients of $\left [ 0.1, 0.3 \right ]$, and denoise for 128, 500, or 1000 steps.
Figure~\ref{fig:fid_cas_log_var_256x} shows Pareto curves of FID vs.\ CAS for the $64\!\times\!64 \rightarrow 256\!\times\!256$ super-resolution module across different noise conditioning augmentation values using a guidance weight of 1.0.
These curves demonstrate the combined impact of the log-variance mixing coefficient and condition noise augmentation in achieving an optimal balance between FID and CAS.

Overall, the results suggest that FID and CAS are highly correlated, with smaller guidance weights leading to better CAS but negatively affecting Inception Score. 
We observe that using noise augmentation of 0 yields the lowest FID score for all values of guidance weights for super-resolution models. 
Nevertheless, it is worth noting that while larger amounts of noise augmentation tend to increase FID, they also produce more diverse samples, as also observed by \cite{saharia2022photorealistic}. Results of these studies are available in the Appendix.  

Based on these sweeps, taking FID and CAS into account, we selected guidance of 1.25 when sampling from the base model,
and 1.0 for other resolutions. 
We use DDPM sampler \cite{ho2020denoising} log-variance mixing coefficients of 0.0 and 0.1 for $64\times64$ samples and $256\times256$ samples respectively, with 1000 denoising steps. 
At resolution $1024\times1024$ we use a DDIM sampler \cite{song2020denoising} with 32 steps, as in \cite{saharia2022photorealistic}. 
We do not use noise conditioning augmentation for sampling.

\subsection{Generation Protocol}
\vspace*{-0.1cm}

We use the fine-tuned Imagen model with the optimized sampling  hyper-parameters to generate synthetic data resembling the training split of ImageNet dataset. 
This means that we aim to produce the same quantity of images for each class as found in the real ImageNet dataset while keeping the same class balance as the original dataset.
We then constructed large-scale training datasets with ranging from 1.2M to 12M images, i.e., between $1\times$ to $10\times$ the size of the original ImageNet training set.

%% file: results.tex
\section{Results}
\vspace*{-0.1cm}

\begin{table*}[t]
\centering
\begin{tabular}{lccc}
\toprule 

Model                                                                           & FID train             & FID validation    & IS                   \\ \toprule
64x64 resolution                                                                & \multicolumn{1}{l}{}   & \multicolumn{1}{l}{} & \multicolumn{1}{l}{} \\ \cmidrule{1-1}
BigGAN-deep  (Dhariwal \& Nichol, 2021)~\cite{dhariwal2021diffusion}            & 4.06                   & -                & -                    \\
Improved DDPM (Nichol \& Dhariwal, 2021)~\cite{nichol2021improved}              & 2.92                   & -                & -                    \\
ADM (Dhariwal \& Nichol, 2021)~\cite{dhariwal2021diffusion}                     & 2.07                   & -                & -                    \\
CDM (Ho et al, 2022)~\cite{ho2022cascaded}                                      & 1.48                   & 2.48             & 67.95 ± 1.97         \\
RIN  (Jabri et al., 2022)~\cite{jabri2022scalable}                              & 1.23                   & -                & 66.5                 \\
RIN + noise schedule (Chen, 2023)~\cite{chen2023importance}                     & 2.04                   & -                & 55.8                 \\
{\bf Ours} (Fine-tuned Imagen)                                                  & 1.21                   & 2.51             & 85.77 ± 0.06         \\ \midrule

256x256 resolution                                                              & \multicolumn{1}{l}{} & \multicolumn{1}{l}{} & \multicolumn{1}{l}{} \\ \cmidrule{1-1}
BigGAN-deep (Brock et al., 2019)~\cite{brock2019large}                          & 6.9                  & -                  & 171.4 ± 2.00         \\
VQ-VAE-2 (Razavi et al., 2019)~\cite{razavi2019generating}                      & 31.11                & -                  & -                    \\
SR3 (Saharia et al., 2021)~\cite{saharia2022palette}                            & 11.30                & -                  & -                    \\
LDM-4 (Rombach et al., 2022)~\cite{rombach2022high}                             & 10.56                & -                  & 103.49               \\
DiT-XL/2 (Peebles \& Xie, 2022)~\cite{peebles2022gan}                           & 9.62                 & -                  & 121.5                \\
ADM (Dhariwal \& Nichol, 2021)~\cite{dhariwal2021diffusion}                     & 10.94                & -                  & 100.98               \\
ADM+upsampling (Dhariwal \& Nichol, 2021)~\cite{dhariwal2021diffusion}          & 7.49                 & -                  & 127.49               \\
CDM (Ho et al, 2022)~\cite{ho2022cascaded}                                      & 4.88                 & 3.76               & 158.71 ± 2.26        \\
RIN (Jabri et al., 2022)~\cite{jabri2022scalable}                               & 4.51                 & 4.51               & 161.0                \\
RIN + noise schedule (Chen, 2023)~\cite{chen2023importance}                     & 3.52                 & -                  & 186.2                \\
Simple Diffusion (U-Net) (Hoogeboom et al., 2023)~\cite{hoogeboom2023simple}    & 3.76                 & 2.88               & 171.6 ± 3.07         \\
Simple Diffusion (U-ViT L) (Hoogeboom et al., 2023)~\cite{hoogeboom2023simple}  & 2.77                 & 3.23               & 211.8 ± 2.93         \\
{\bf Ours}  (Fine-tuned Imagen)                                                 & 1.76                 & 2.81               & 239.18 ± 1.14        \\ 

\bottomrule
\end{tabular}
\vspace*{0.15cm}
\caption{
Comparison of sample quality of synthetic ImageNet datasets measured by FID and Inception Score (IS) between our fine-tuned Imagen model and generative models in the literature. We achieve SOTA FID and IS on ImageNet generation among other existing models, including class-conditional and guidance-based sampling without any design changes.
}
\label{tab:fid-is-literature}
\vspace*{-0.05cm}
\end{table*}

\subsection{Sample Quality: FID and IS}
\vspace*{-0.1cm}

Despite the shortcomings described in Sec.~\ref{sec:background}, FID~\cite{heusel2017gans} and Inception Score~\cite{salimans2016improved} remain standard metrics for evaluating generative models. 
Table~\ref{tab:fid-is-literature} reports FID and IS for our approach and existing class-conditional and guidance-based approaches.
Our fine-tuned model outperforms all the existing methods, including state-of-the-art methods that use U-Nets~\cite{ho2022cascaded} and larger U-ViT models trained solely on ImageNet data~\cite{hoogeboom2023simple}.  This suggests that large-scale pretraining followed by fine-tuning on domain-specific target data is an effective strategy to achieve better visual quality with diffusion models, as  measured by FID and IS.
Figure \ref{fig:samples} shows imaage samples from the fine-tuned model (see Appendix for more).
Note that our state-of-the-art FID and IS on ImageNet are obtained without any design changes, i.e., by simply adapting an off-the-shelf, diffusion-based text-to-image model to new data through fine-tuning. This is a promising result indicating that in a resource-limited setting, one can improve the performance of diffusion models by fine-tuning model weights and adjusting sampling parameters.

\begin{figure*}[]
\begin{center}
\includegraphics[width=0.33\textwidth]{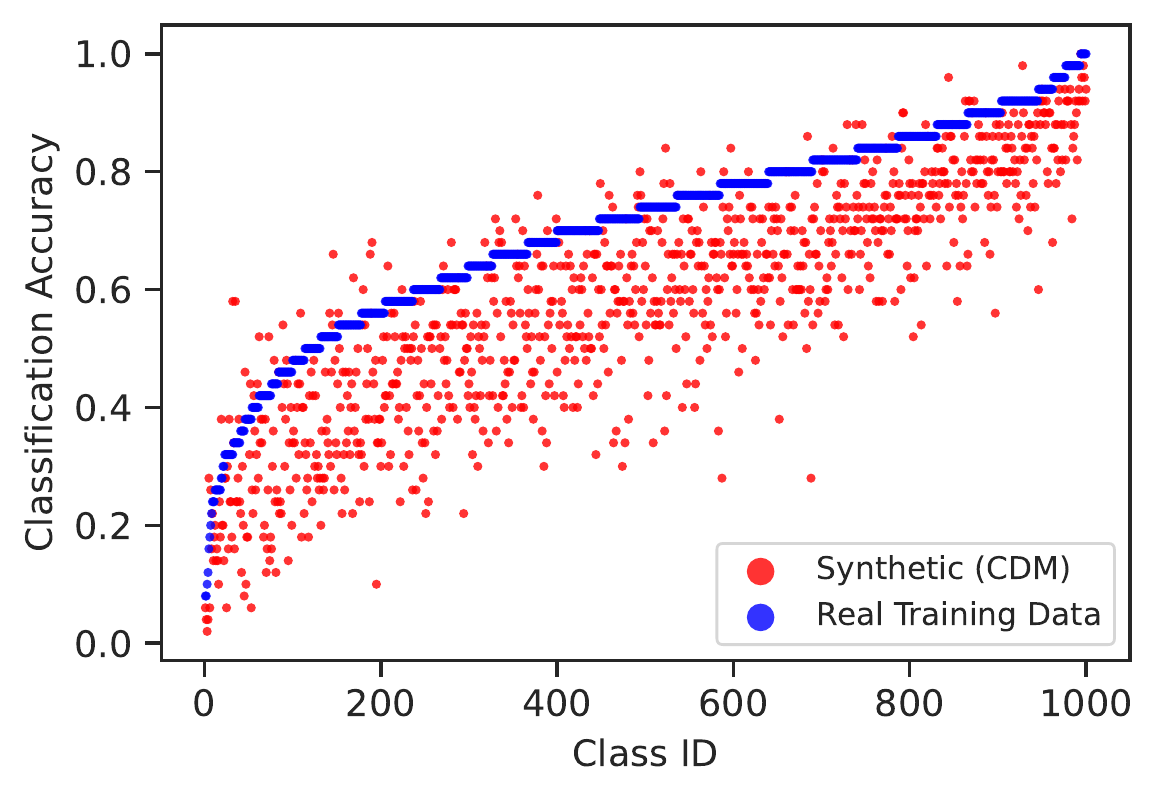}
\includegraphics[width=0.33\textwidth]{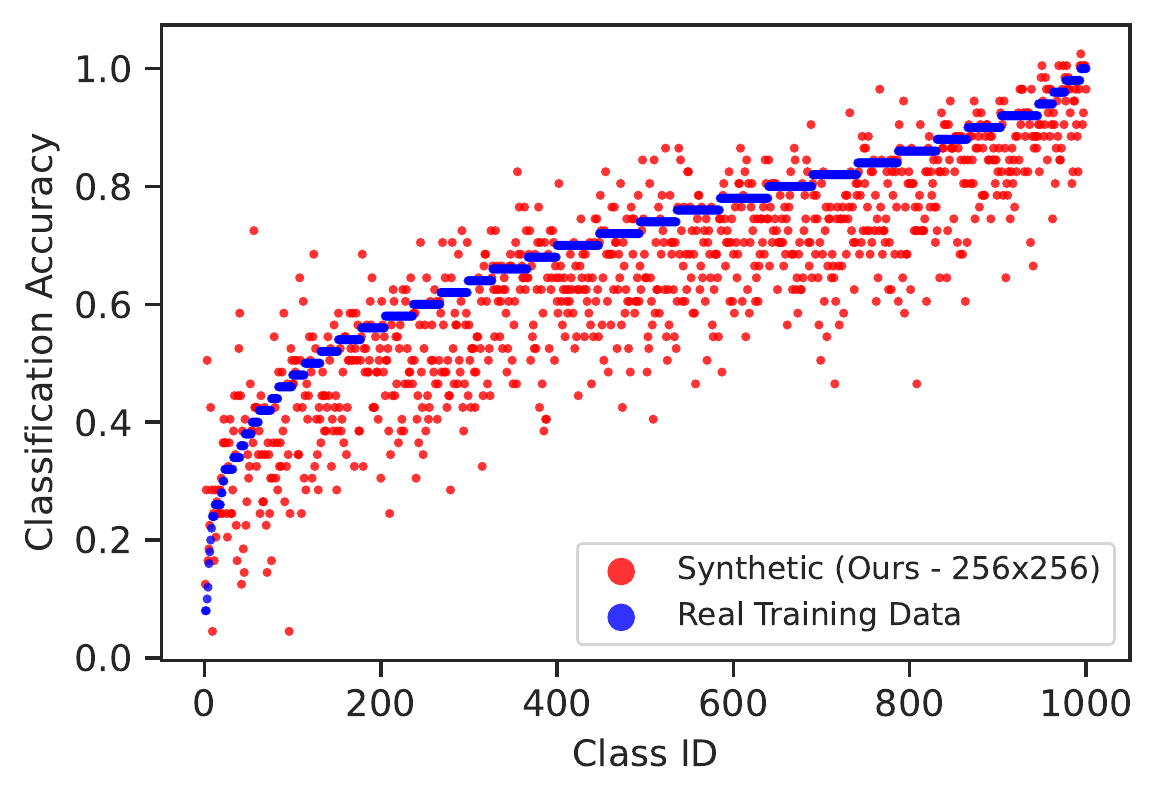} 
\includegraphics[width=0.33\textwidth]{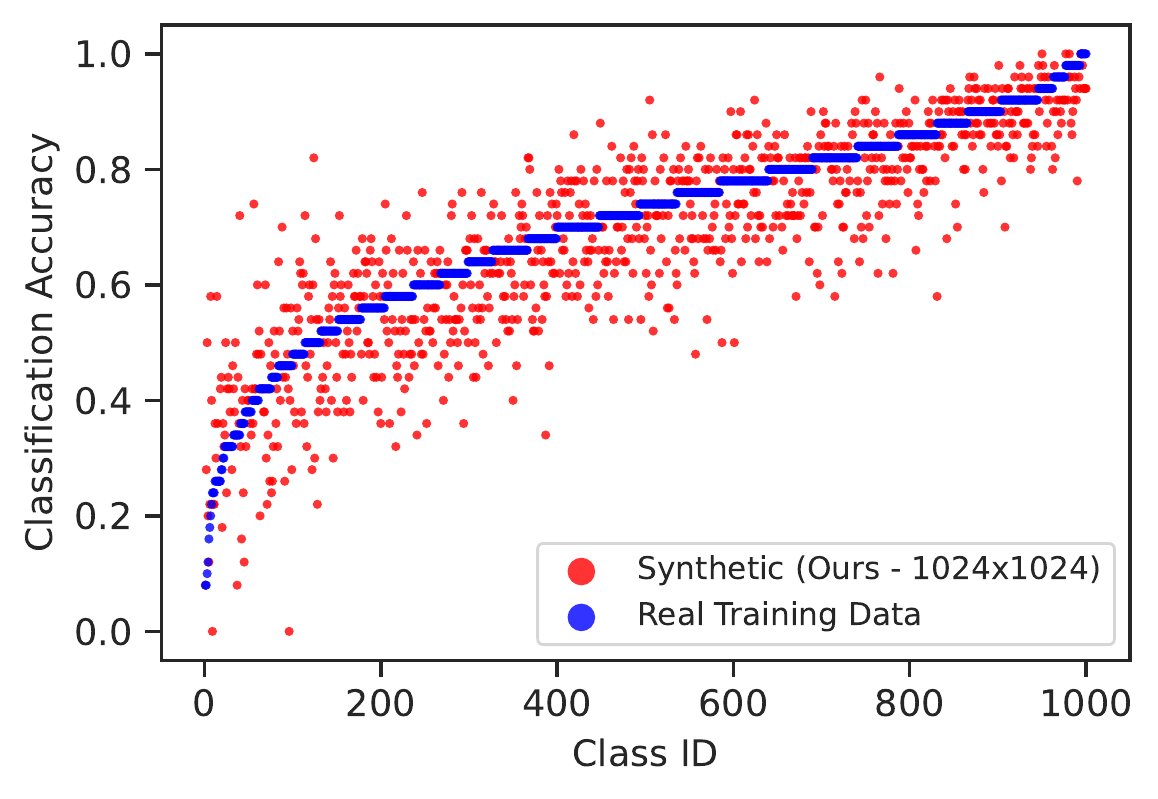}
\end{center}
\vspace*{-0.5cm}
\caption{Class-wise classification accuracy comparison accuracy of models trained on real data (blue) and generated data (red). 
{\bf Left}: The $256 \times 256$ CDM  model \cite{ho2022cascaded}. {\bf Middle and right:} Our fine-tuned Imagen model at $256 \times 256$ and $1024 \times 1024$.
}
\label{fig:per-class}
\vspace*{-0.0cm}
\end{figure*}

\begin{table*}
\centering
\vspace*{0.2cm}
\begin{tabular}{lcc}
\toprule
Model                                                       & Top-1 Accuracy (\%) & Top-5 Accuracy(\%) \\ \toprule
Real                                                        & 73.09        & 91.47        \\ \midrule
BigGAN-deep (Brock et al., 2019)~\cite{brock2019large}      & 42.65        & 65.92        \\
VQ-VAE-2 (Razavi et al, 2019)~\cite{razavi2019generating}   & 54.83        & 77.59        \\
CDM (Ho et al,  2022)~\cite{ho2022cascaded}                 & 63.02        & 84.06        \\ \midrule
{\bf Ours}  (256$\times$256 resolution)                     & 64.96        & 85.66        \\
{\bf Ours}  (1024$\times$1024 resolution)                   & 69.24        & 88.10        \\
\bottomrule       
\end{tabular}
\vspace*{0.1cm}
\caption{
Classification Accuracy Scores (CAS) for $256\!\times\! 256$ and $1024\!\times\! 1024$ generated samples. CAS for real data and other models are obtained from~\cite{ravuri2019classification} and~\cite{ho2022cascaded}. Our results indicate that the fine-tuned generative diffusion model outperforms the previous methods by a substantial margin.
}
\label{tab:cas-results}
\vspace*{0.1cm}
\end{table*}

\subsection{Classification Accuracy Score}
\vspace*{-0.1cm}

As noted above, classification accuracy score (CAS)~\cite{ravuri2019classification} is a better proxy than FID and IS for performance of downstream training on generated data. 
CAS measures ImageNet classification accuracy on the real test data for a model trained solely on synthetic samples.
In keeping with the CAS protocol~\cite{ravuri2019classification}, we train a standard ResNet-50 architecture on a single crop from each training image. Models are trained for 90 epochs with a batch size of 1024 using SGD with momentum (see Appendix \ref{app:hyperparameters} for details).
Regardless of the resolution of the generated data, for CAS training and evaluation, we resize images to $256\! \times\! 256$ (or, for real images, to 256 pixels on the shorter side) and then take a $224\! \times\! 224$ pixel center crop.

Table~\ref{tab:cas-results} reports CAS for samples from our fine-tuned models at resolutions $256\! \times\! 256$ and $1024\! \times\! 1024$. CAS for real data and for other models are taken from~\cite{ravuri2019classification} and~\cite{ho2022cascaded}. 
The results indicate that our fine-tuned class-conditional models outperform the previous methods in the literature at  $256 \times 256$ resolution by a good margin, for both Top-1 and Top-5 accuracy.
Interestingly, results are markedly better for $1024\times 1024$ samples, even though these samples are down-sampled to $256\times 256$ during classifier training.
As reported in Table~\ref{tab:cas-results},
we achieve the SOTA Top-1 classification accuracy score of 69.24\% at resolution $1024\! \times\! 1024$. This greatly narrows the gap with the ResNet-50 model trained on real data.

Figure \ref{fig:per-class} shows the accuracy of models trained on generative data (red) compared to a model trained on real data (blue) for each of the 1000 ImageNet classes
(cf.\ Fig.\ 2 in \cite{ravuri2019classification}).
From Figure \ref{fig:per-class} (left) one can see that the ResNet-50 trained on CDM samples is weaker than the model trained on real data, as most red points fall below the blue points.
For our fine-tuned Imagen models (Figure \ref{fig:per-class} middle and right), however, there are more classes for which the models trained on generated data outperform the model trained on real data.
This is particularly clear at $1024 \!\times\!1024$.

\subsection{Classification Accuracy with Different Models}
\vspace*{-0.1cm}

To further evaluate the discriminative power of the synthetic data, compared to the real ImageNet-1K data, we analyze the classification accuracy of models with different architectures, input resolutions, and model capacities. 
We consider multiple ResNet-based and Vision Transformers (ViT)-based~\cite{dosovitskiy2020image} classifiers including ResNet-50~\cite{he2016deep}, ResNet-RS-50, ResNet-RS-152x2, ResNet-RS-350x2~\cite{bello2021revisiting}, ViT-S/16~\cite{beyer2022better}, and DeiT-B~\cite{touvron2021training}. 
The models trained on real, synthetic, and the combination of real and synthetic data are all trained in the same way, consistent with the training recipes specified by authors of these models on ImageNet-1K, and our  results on real data agree with the published results. The Appendix has more details on model training.

Table \ref{tab:data-augmentation} reports the Top-1 validation accuracy of multiple ConvNet and Transformer models when trained with the 1.2M real ImageNet training images, with 1.2M generated images, and when the generative samples are used to augment the real data.
As one might expect, models trained solely on generated samples perform worse than models trained on real data.
Nevertheless, augmenting real data with synthetic images from the diffusion model yields a substantial boost in performance across all   classifiers tested.

\subsection{Merging Real and Synthetic Data at Scale}
\vspace*{-0.1cm}

We next consider how performance of a ResNet-50 classifier depends on the amount of generated data that is used to augment the real data. Here we follow conventional training recipe and train with random crops for 130 epochs, resulting in a higher ResNet-50 accuracy here than in the CAS results in Table \ref{tab:cas-results}. 
The Appendix provides  training details.

Ravuri and Vinyals \cite{ravuri2019classification} (Fig.\ 5) found that for almost all models tested, mixing generated samples with real data degrades Top-5 classifier accuracy.
For Big-GAN-deep \cite{brock2019large} with low truncation values (sacrificing diversity for sample quality), accuracy increases marginally with small amounts of generated data, but then drops below models trained solely on real data when the amount of generated data approaches the size of the real train set. 

\begin{table*}[t]
\centering
\begin{tabular}{l|c|c|cccc} 
\toprule
Model  & Input Size & Params (M)    & Real Only & Generated Only &  Real + Generated & Performance $\Delta$    \\ \toprule
\multicolumn{7}{c}{ConvNets}   \\ \toprule
ResNet-50      & 224$\times$224 & 36   & 76.39  & 69.24   & 78.17           & \textcolor{black}{+1.78} \\
ResNet-101     & 224$\times$224 & 45   & 78.15  & 71.31   & 79.74           & \textcolor{black}{+1.59} \\
ResNet-152     & 224$\times$224 & 64   & 78.59  & 72.38   & 80.15           & \textcolor{black}{+1.56} \\ \midrule
ResNet-RS-50   & 160$\times$160 & 36   & 79.10  & 70.72   & 79.97           & \textcolor{black}{+0.87} \\
ResNet-RS-101  & 160$\times$160 & 64   & 80.11  & 72.73   & 80.89           & \textcolor{black}{+0.78} \\
ResNet-RS-101  & 190$\times$190 & 64   & 81.29  & 73.63   & 81.80           & \textcolor{black}{+0.51} \\
ResNet-RS-152  & 224$\times$224 & 87   & 82.81  & 74.46   & 83.10           & \textcolor{black}{+0.29} \\ \toprule
\multicolumn{7}{c}{Transformers}   \\ \toprule
ViT-S/16      & 224$\times$224 & 22   & 79.89  & 71.88   & 81.00          & \textcolor{black}{+1.11}  \\
DeiT-S        & 224$\times$224 & 22   & 78.97  & 72.26   & 80.49          & \textcolor{black}{+1.52}  \\
DeiT-B        & 224$\times$224 & 87   & 81.79  & 74.55   & 82.84          & \textcolor{black}{+1.04}  \\
DeiT-B        & 384$\times$384 & 87   & 83.16  & 75.45   & 83.75          & \textcolor{black}{+0.59}  \\
DeiT-L        & 224$\times$224 & 307  & 82.22  & 74.60   & 83.05          & \textcolor{black}{+0.83}  \\
\bottomrule
\end{tabular}
\vspace*{-0.05cm}
\caption{Comparison of classifier Top-1 Accuracy (\%) performance when 1.2M generated images are used for generative data augmentation.
Models trained solely on generated samples perform worse than models trained on real data.
Nevertheless, augmenting the real data with data generated from the fine-tuned diffusion model provides a substantial boost in performance across many different classifiers.} 
\label{tab:data-augmentation}
\vspace*{-0.05cm}
\end{table*}

\begin{table}[t]
\centering
\begin{tabular}{c|cc}
\toprule
Train Set (M)  & 256$\times$256        & 1024$\times$1024   \\ \toprule
1.2 & 76.39 ± 0.21  & 76.39 ± 0.21 \\ \hline
2.4  & 77.61 ± 0.08 \scriptsize{\textcolor{ourdarkgreen}{(+1.22)}}  & 78.12 ± 0.05 \scriptsize{\textcolor{ourdarkgreen}{(+1.73)}} \\ 
3.6  & 77.16 ± 0.04 \scriptsize{\textcolor{ourdarkgreen}{(+0.77)}}  & 77.48 ± 0.04 \scriptsize{\textcolor{ourdarkgreen}{(+1.09)}} \\
4.8  & 76.52 ± 0.04 \scriptsize{\textcolor{ourdarkgreen}{(+0.13)}}  & 76.75 ± 0.07 \scriptsize{\textcolor{ourdarkgreen}{(+0.36)}} \\
6.0  & 76.09 ± 0.08 \scriptsize{\textcolor{ourdarkred}{(-0.30)}}    & 76.34 ± 0.13 \scriptsize{\textcolor{ourdarkred}{(-0.05)}} \\
7.2  & 75.81 ± 0.08 \scriptsize{\textcolor{ourdarkred}{(-0.58)}}    & 75.87 ± 0.09 \scriptsize{\textcolor{ourdarkred}{(-0.52)}} \\
8.4  & 75.44 ± 0.06 \scriptsize{\textcolor{ourdarkred}{(-0.95)}}    & 75.49 ± 0.07 \scriptsize{\textcolor{ourdarkred}{(-0.90)}} \\
9.6  & 75.28 ± 0.10 \scriptsize{\textcolor{ourdarkred}{(-1.11)}}    & 74.72 ± 0.20 \scriptsize{\textcolor{ourdarkred}{(-1.67)}} \\
10.8 & 75.11 ± 0.12 \scriptsize{\textcolor{ourdarkred}{(-1.28)}}    & 74.14 ± 0.13 \scriptsize{\textcolor{ourdarkred}{(-2.25)}} \\
12.0 & 75.04 ± 0.05 \scriptsize{\textcolor{ourdarkred}{(-1.35)}}    & 73.70 ± 0.09 \scriptsize{\textcolor{ourdarkred}{(-2.69)}} \\ 
\bottomrule
\end{tabular}
\vspace*{-0.00cm}
\caption{Scaling the training dataset by adding synthetic images, at resolutions
$256\!\times\! 256$ and $1024\!\times\! 1024$.
The baseline Top-1 accuracy of the classifier trained on real data is $76.39\pm0.21$. The number in parenthesis shows the change obtained over baseline with the addition of generated data.
}
\label{tab:scaling-gen-data}
\vspace*{-0.00cm}
\end{table}

\begin{figure}[t]
\vspace*{-0.1cm}
\begin{center}
\includegraphics[width=0.34\textwidth]{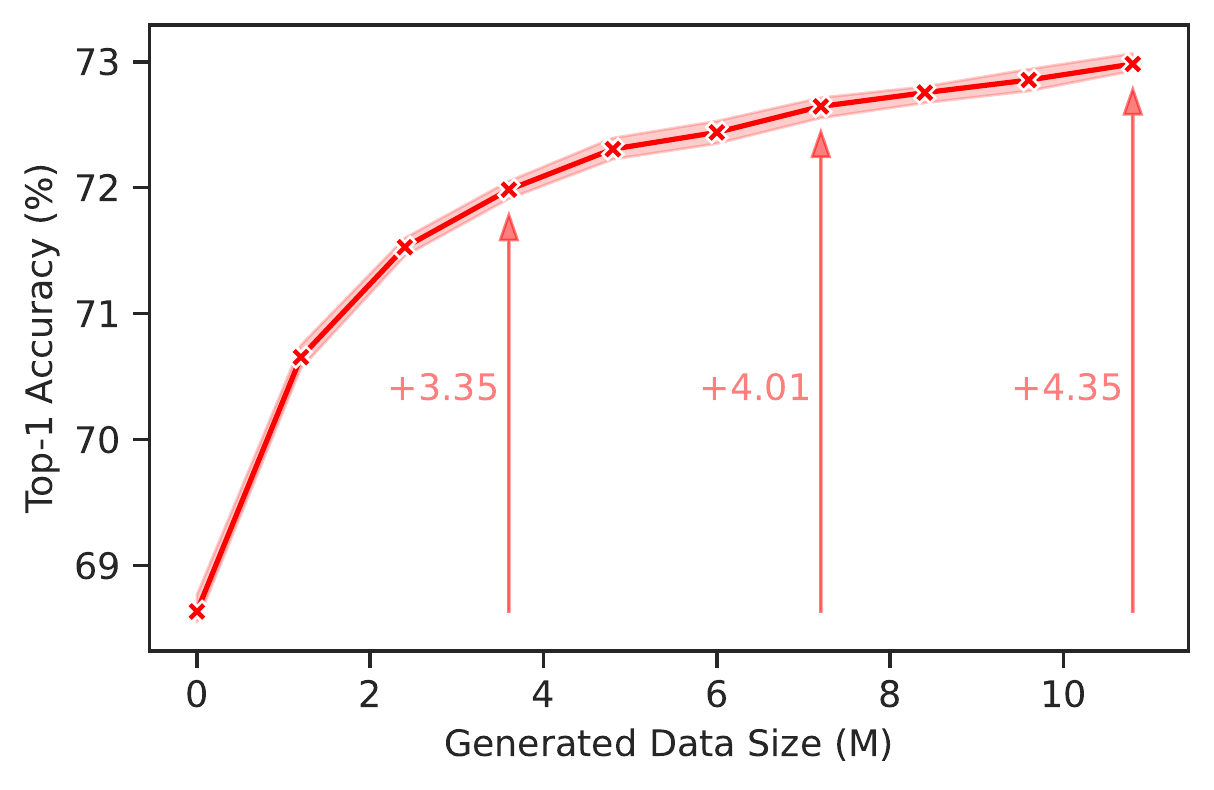}
\end{center}
\vspace*{-0.1cm}
\caption{Improved classification accuracy of ResNet-50 with increasing numbers of synthetic images added to real training data at resolution 
$64\!\times\! 64$.
}
\label{fig:scaling-64A}
\vspace*{-0.1cm}
\end{figure}

Figure \ref{fig:scaling-64A} shows that, for $64\times 64$ images, performance continues to improve as the amount of generated data increases up to nine times the amount of real data, to a total dataset size of 12M images.
Performance with higher resolution images, however, does not continue to improve with similarly large amounts of generative data augmentation.
Table \ref{tab:scaling-gen-data} reports performance as the amount of generated data increased over the same range, up to $9\times$ the amount of real data, at resolutions $256\!\times\! 256$ and $1024\!\times\! 1024$.
The performance boost remains significant with fine-tuned diffusion models for synthetic data up to a factor of 4 or
5 times the size of the real ImageNet training set, a significant improvement over results reported in \cite{ravuri2019classification}.

%% file: conclusion.tex
\vspace*{-0.1cm}
\section{Conclusion}
\vspace*{-0.1cm}

This paper asks to what extent generative data augmentation is effective with current diffusion models.
We do so in the context of ImageNet classification, a challenging domain as it is extensively explored with highly tuned architectures and training recipes.
Here we show that large-scale text-to-image diffusion models can be fine-tuned to produce class-conditional models with SOTA FID (1.76 at $256\!\times\! 256$ resolution) and Inception Score (239 at $256\!\times\! 256$).
The resulting generative model also yields a new SOTA in Classification Accuracy Scores (64.96 for $256\!\times\! 256$ models, improving to 69.24 for $1024\!\times\! 1024$ generated samples).
And we have shown improvements to ImageNet classification accuracy
extend to large amounts of generated data, across a range of ResNet and Transformer-based models.

While these results are encouraging, many questions remain. One concerns the boost in CAS at resolution $1024 \!\times\!1024$, suggesting that the larger images capture more useful image structure than those at $256 \!\times\!256$, even though the $1024 \!\times\!1024$ images are downsampled to $256 \!\times\!256$ before being center-cropped to $224\!\times\!224$ for input to ResNet-50. 
Another concerns the sustained gains in classification accuracy with large amounts of synthetic data at $64\!\times\!64$ (Fig.\ \ref{fig:scaling-64A}); there is less information at low resolutions for training, and hence a greater opportunity for augmentation with synthetic images.
At high resolutions (Tab.\ \ref{tab:scaling-gen-data}) performance drops for synthetic datasets larger than 1M images, which may indicate bias in the generative model, and the need for more sophisticated training methods with synthetic data. These issues remain topics of on-going research.

\subsection*{Acknowledgments}
\vspace*{-0.1cm}
We thank Jason Baldridge and Ting Chen for their valuable feedback. We also extend thanks to William Chan, Saurabh Saxena, and Lala Li for helpful discussions, feedback, and their support with the Imagen code. 

\vspace*{-0.1cm}

%% file: appendix.tex
\onecolumn

\renewcommand{\thesection}{A.\arabic{section}}
\renewcommand{\thefigure}{A.\arabic{figure}}
\renewcommand{\thetable}{A.\arabic{table}} 
\renewcommand{\theequation}{A.\arabic{equation}} 

\setcounter{section}{0}
\setcounter{figure}{0}
\setcounter{table}{0}
\setcounter{equation}{0}

\begin{center}
\noindent \textbf{\LARGE{Appendix}}\\
\end{center}
\normalfont

\section{Hyper-parameters for Imagen fine-tuning and sample generation.}
\label{app:Imagen-HPs}

The quality, diversity, and speed of text-conditioned diffusion model sampling are strongly affected by multiple hyper-parameters.
These include the number of diffusion steps, where larger numbers of diffusion steps are often associated with higher quality images and lower FID.
Another hyper-parameter is the amount of noise-conditioning augmentation \cite{saharia2022photorealistic}, which adds Gaussian noise to the output of one stage of the Imagen cascade at training time, prior to it being input to the subsequent super-resolution stage. 
We considered noise levels between 0 and 0.5
(with images in the range [0,1]), where adding more noise during training degrades more fine-scale structure, thereby forcing the subsequent super-resolution stage to be more robust to variability in the images generated from the previous stage.

During sampling, we use classifier-free guidance \cite{ho2022classifier,nichol2021glide}, but with smaller guidance weights than Imagen, favoring diversity over image fidelity to some degree.
With smaller guidance weights, one does not require dynamic thresholding  \cite{saharia2022photorealistic} during inference; instead we opt for a static threshold to clip large pixel values at each step of denoising.
Ho et al.\ \cite{ho2022classifier} identify upper and lower bounds on the predictive variance, $\Sigma_\theta(x_t, t)$, used for sampling at each denoising step.
Following \cite{nichol2021improved} (Eq.\ 15) we use a linear (convex) combination of the log upper and lower bounds, the mixing parameter for which is referred to as the logvar parameter.
Figures \ref{fig:fid_is_cas_pareto_64x} and
\ref{fig:fid_cas_log_var_256x} show the dependence of FID, IS and Classification Accuracy Scores on guidance weight and logvar mixing coefficient for the base model at resolution $64\!\times\!64$ and the $64\! \rightarrow\! 256$ super-resolution model.
These were used to help choose model hyyper-parameters for large-scale sample generation.

Below are further results relate to hyperparameter selection and its impact on model metrics.

\begin{figure}[h!]
\begin{center}
\includegraphics[width=0.32\textwidth]{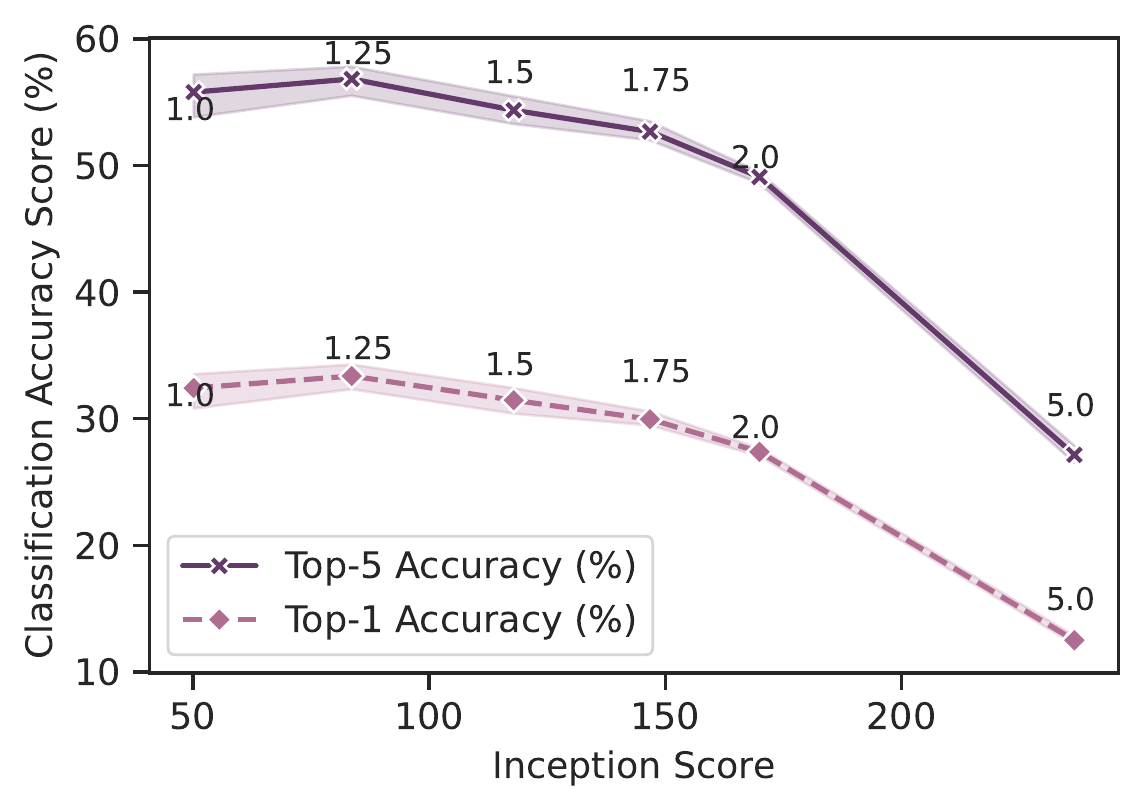}
\includegraphics[width=0.32\textwidth]{figures/fid_is_64x_train.pdf}
\end{center}
\vspace*{-0.25cm}
   \caption{\textbf{Left}: CAS vs IS Pareto curves for train set resolution of $64\!\times\!64$ showing the impact of guidance weights. \textbf{Right}: Train set FID vs IS Pareto curves for resolution of 64x64 showing the impact of guidance weights.}
\label{fig:fid-is-cas-64x}
\end{figure}

\begin{figure}[h!]
\begin{center}
\includegraphics[width=0.32\textwidth]{figures/fid_gw_ddpm_1000_64x_50k.pdf}
\includegraphics[width=0.32\textwidth]{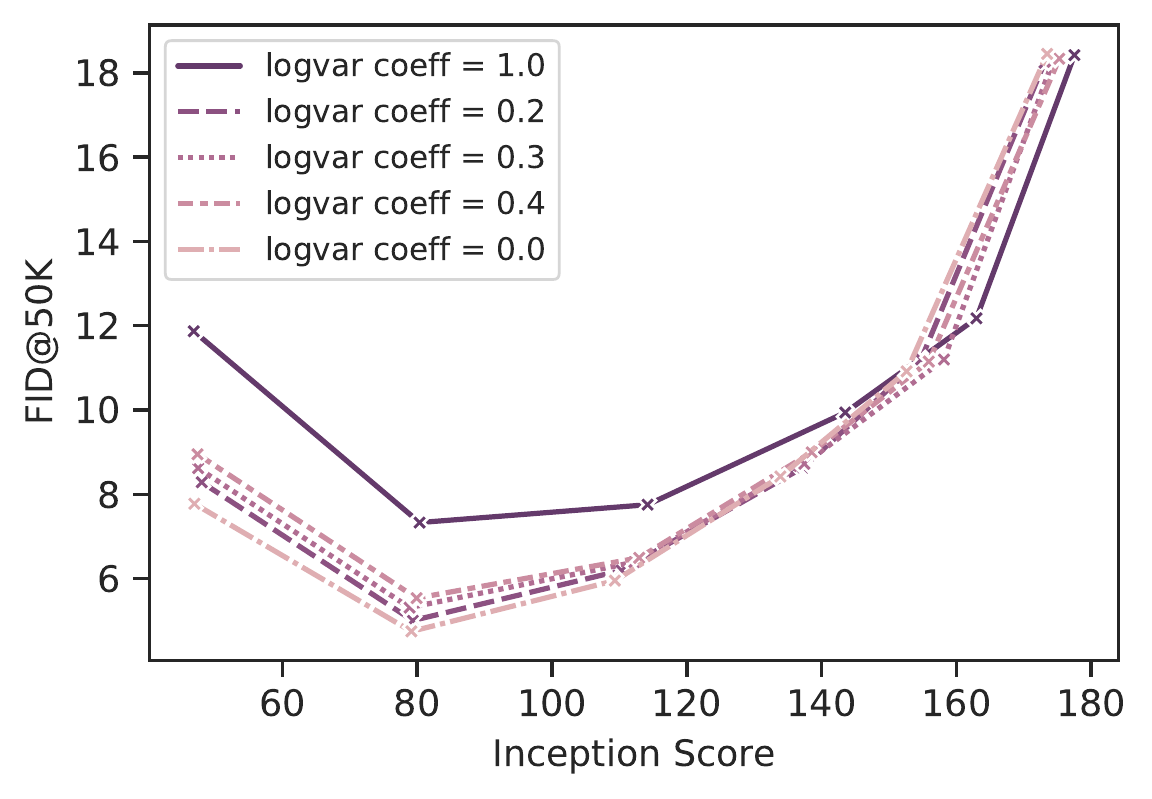}
\end{center}
\vspace*{-0.25cm}
   \caption{Sampling refinement for $64\!\times\!64$ base model. \textbf{Left}: Validation set FID vs.\ guidance weights for different values of log-variance. \textbf{Right}: Validation set FID vs.\ Inception score (IS) when increasing guidance from 1.0 to 5.0. }
\label{fig:fid_ddpm_1000_64x}
\end{figure}

\begin{figure*}[h!]
\begin{center}
\includegraphics[width=0.95\textwidth]{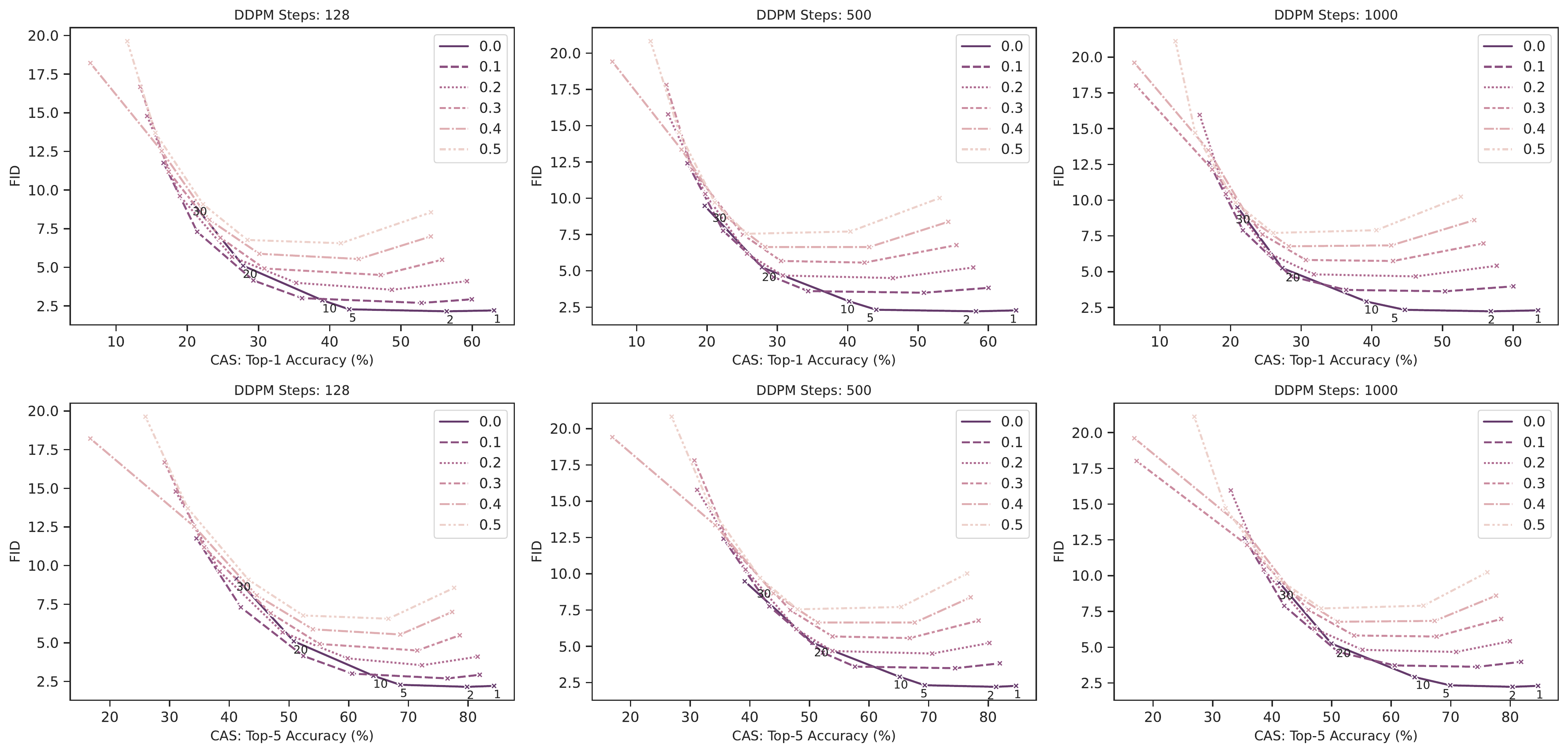}
\end{center}
\vspace*{-0.25cm}
   \caption{Top-1 and Top-5 classification accuracy score (CAS) vs train FID Pareto curves (sweeping over guidance weight) showing the impact of conditioning noise augmentation at $256\!\times\!256$ when sampling with different number of steps. As indicated by number overlaid on each trend line, guidance weight is decreasing from 30 to 1.}
\label{fig:fid_cas_noise_ddpms}
\end{figure*}

\begin{figure*}[h!]
\begin{center}
\includegraphics[width=0.95\textwidth]{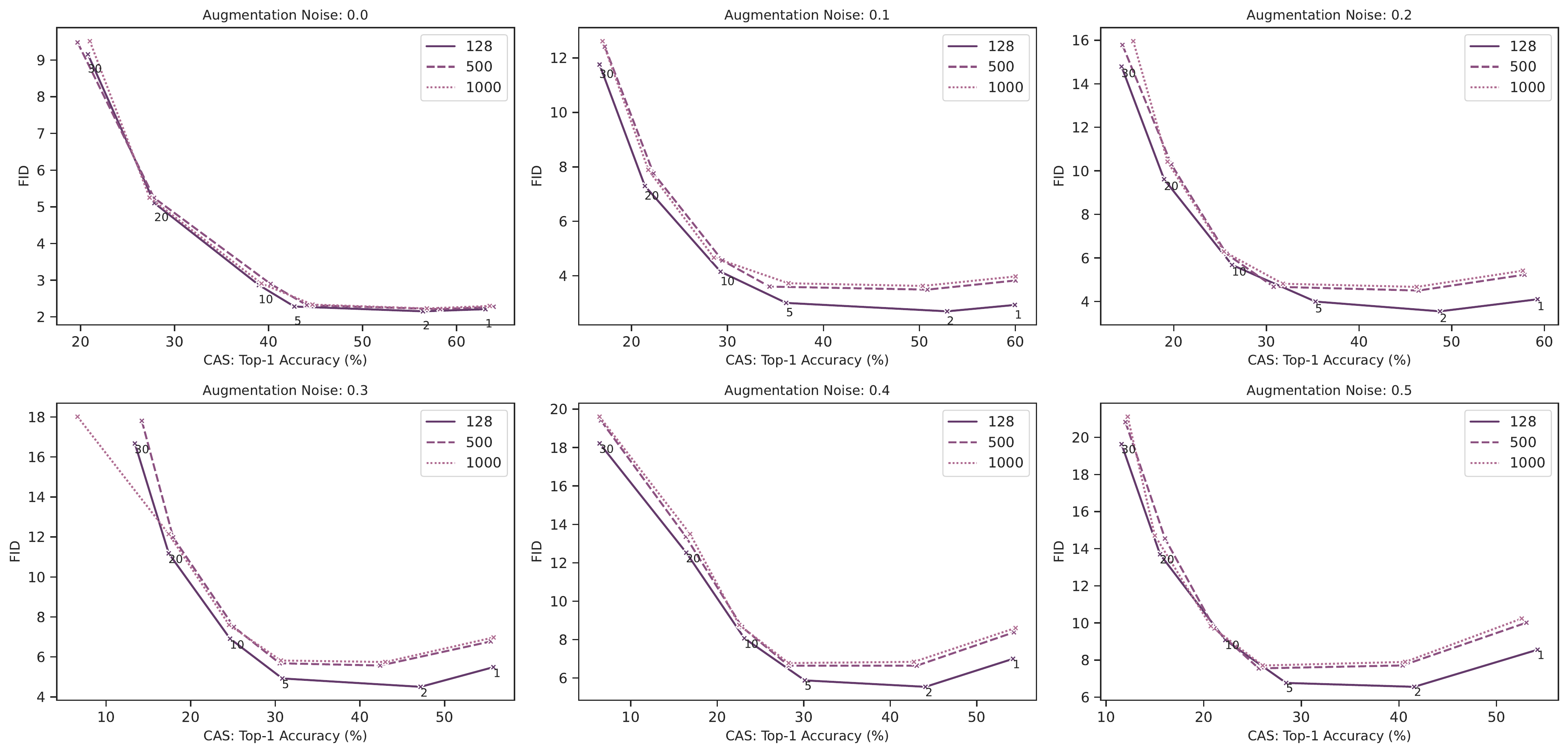}
\end{center}
\vspace*{-0.25cm}
   \caption{Top-1 and Top-5 classification accuracy score (CAS) vs train FID Pareto curves (sweeping over guidance weight) showing the impact of conditioning noise augmentation at $256\!\times\!256$ when sampling with different number of steps at a fixed noise level. As indicated by number overlaid on each trend line guidance weight is decreasing from 30 to 1. At highest noise level (0.5) lowering number sampling step and decreasing guidance can lead to a better joint FID and CAS values. At lowest noise level (0.0) this effect is subtle and increasing sampling steps and lower guidance weight can help to improve CAS. }
\label{fig:fid_cas_noise_steps}
\end{figure*}

\begin{figure}[h!]
\begin{center}
\includegraphics[width=0.32\textwidth]{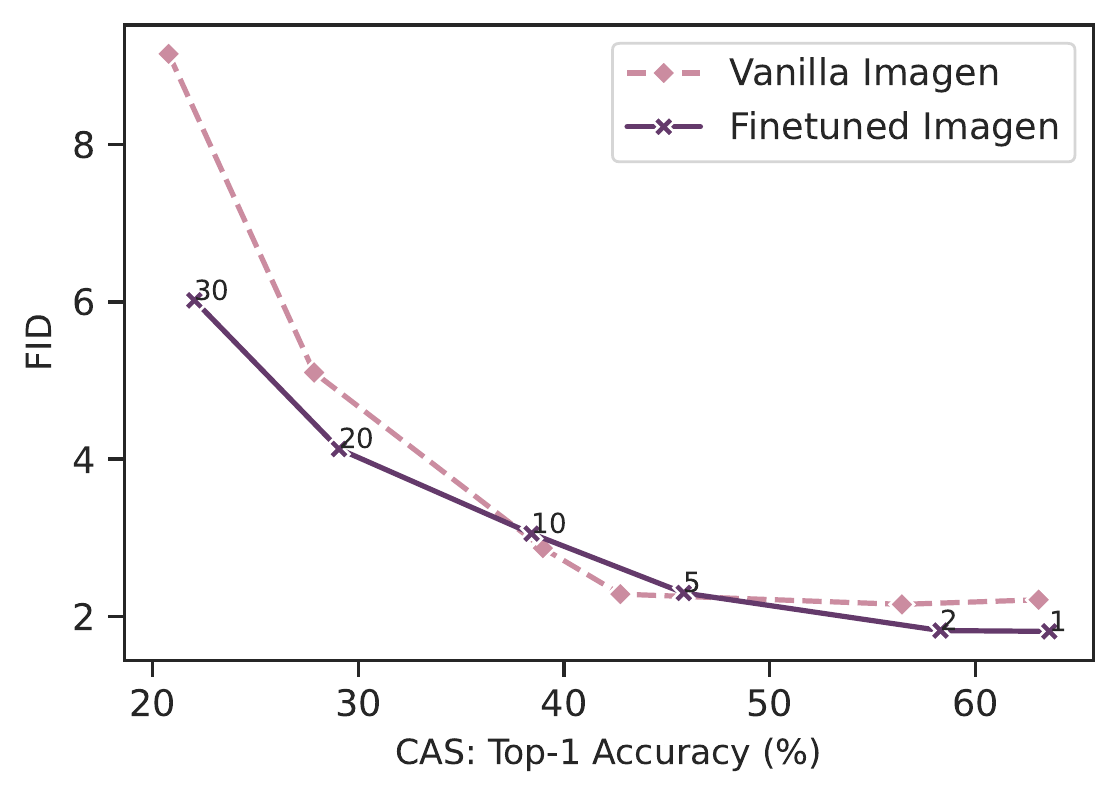}
\includegraphics[width=0.32\textwidth]{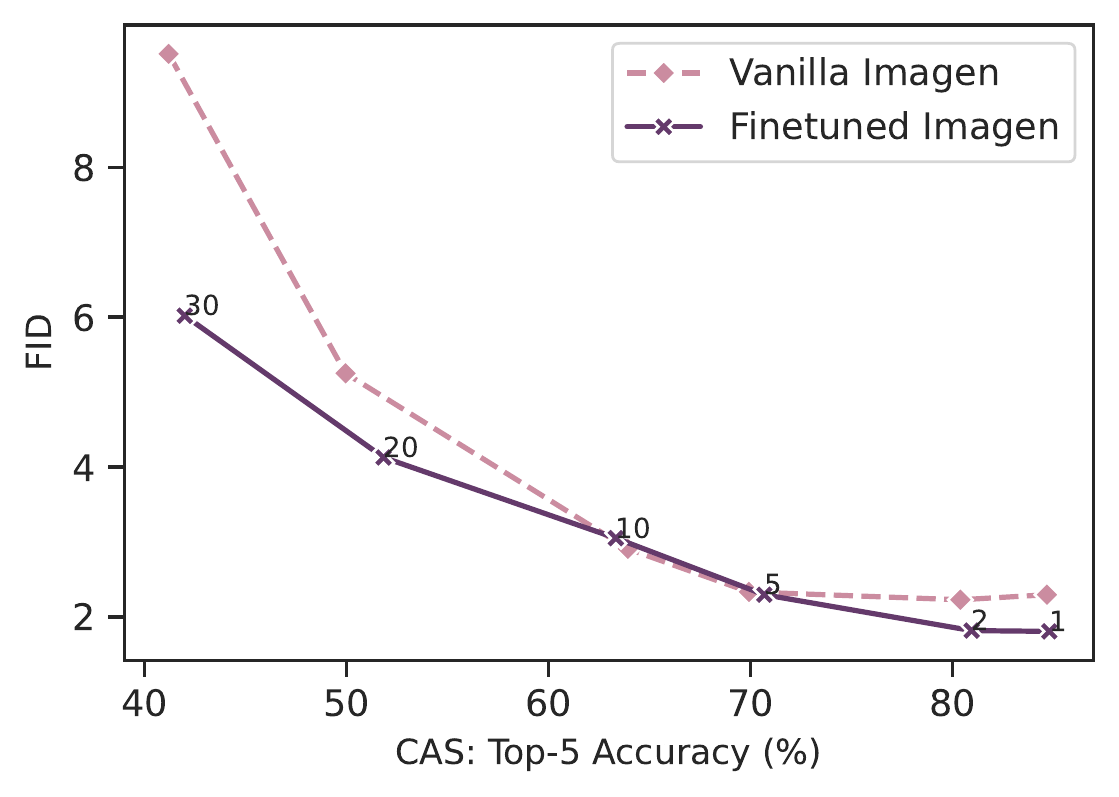}
\end{center}
\vspace*{-0.25cm}
   \caption{Fine-tuning of SR model helps to jointly improve classification accuracy as well as FID  of the vanilla Imagen. }
\label{fig:fid_cas_imagen_vs_finetuned}
\end{figure}

\begin{figure}[h!]
\begin{center}
\includegraphics[width=0.32\textwidth]{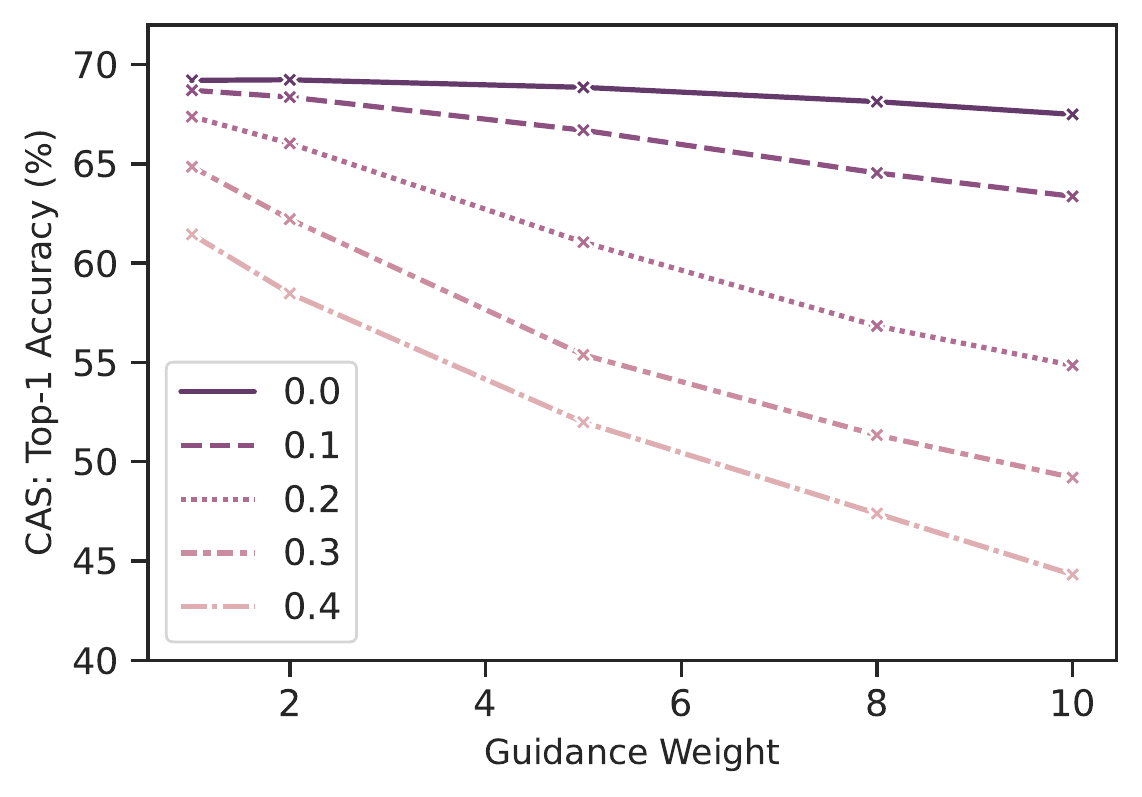}
\includegraphics[width=0.32\textwidth]{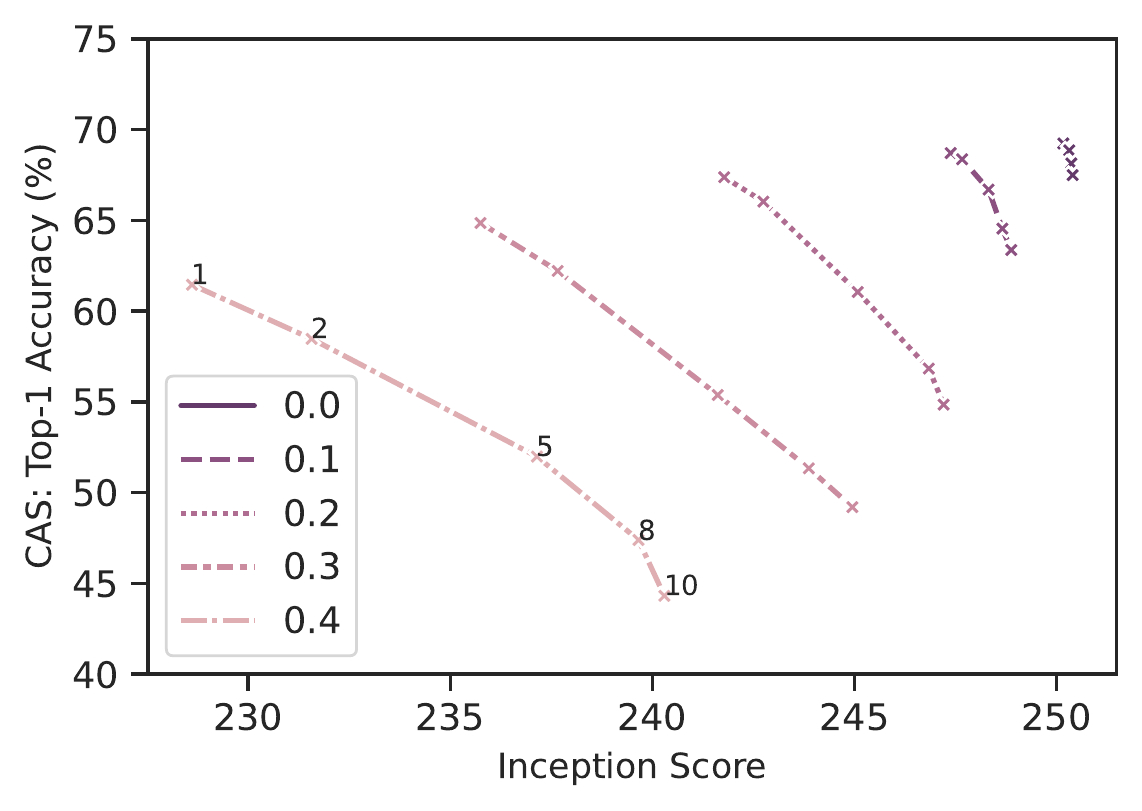}
\end{center}
\vspace*{-0.25cm}
   \caption{Sampling refinement for $1024\!\times\!2014$ super resolution model. \textbf{Left}: CAS vs.\ guidance weights under varying noise conditions. \textbf{Right}: CAS vs.\ Inception score (IS) when increasing guidance from 1.0 to 5.0 under varying noise conditions. }
\label{fig:cas_1024x}
\end{figure}

\begin{figure}[h!]
\begin{center}
\includegraphics[width=0.35\textwidth]{figures/fid_cas_top_1_ddpm_1000.pdf}
\includegraphics[width=0.35\textwidth]{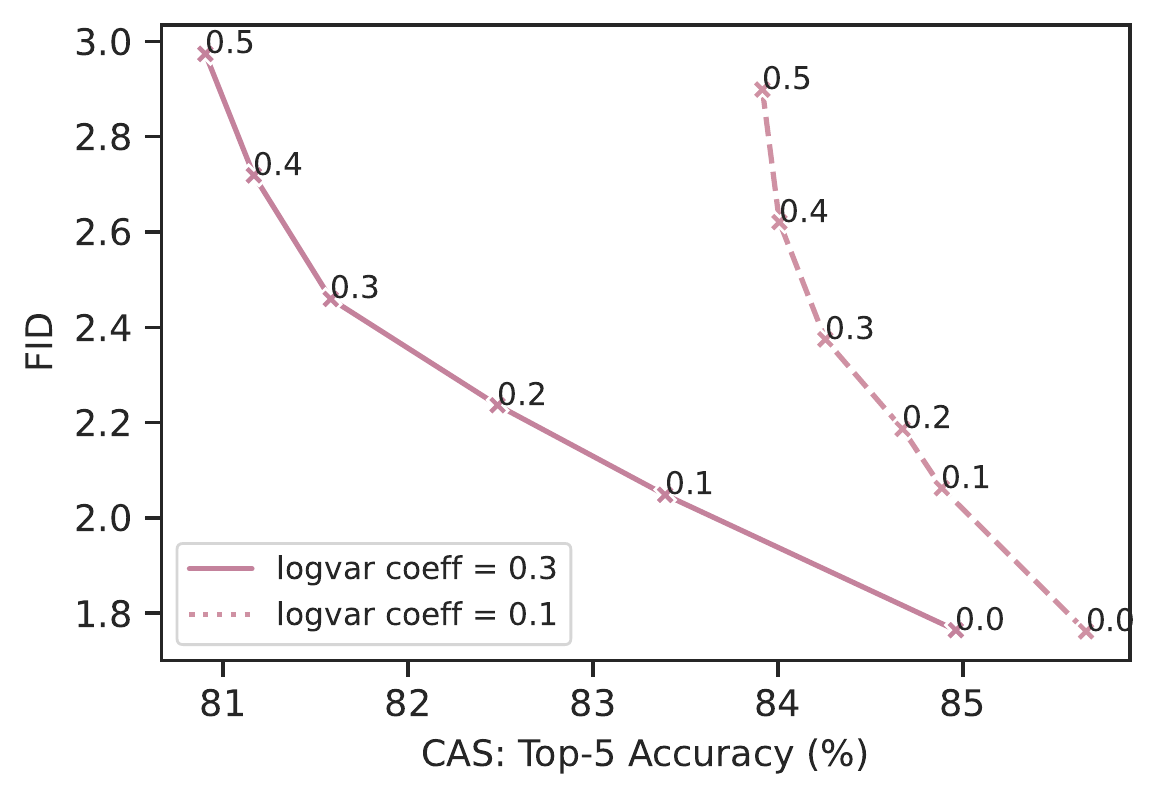}
\end{center}
\vspace*{-0.255cm}
\caption{Training set FID vs.\ classification top-1 and top-5 accuracy Pareto curves under varying noise conditions when the guidance weight is set to 1.0 for resolution $256\!\times\!256$. These curves depict the joint influence of the log-variance mixing coefficient \cite{nichol2021improved} and noise conditioning augmentation \cite{ho2022cascaded} on  FID and CAS.}

\label{fig:fid_cas_pareto_256x}
\end{figure}

\vspace*{5.0cm}

\newpage

\section{Class Alignment of Imagen vs.\ Fine-Tuned Imagen}

\begin{figure}[h!]
\begin{center}
\includegraphics[width=0.81\textwidth]{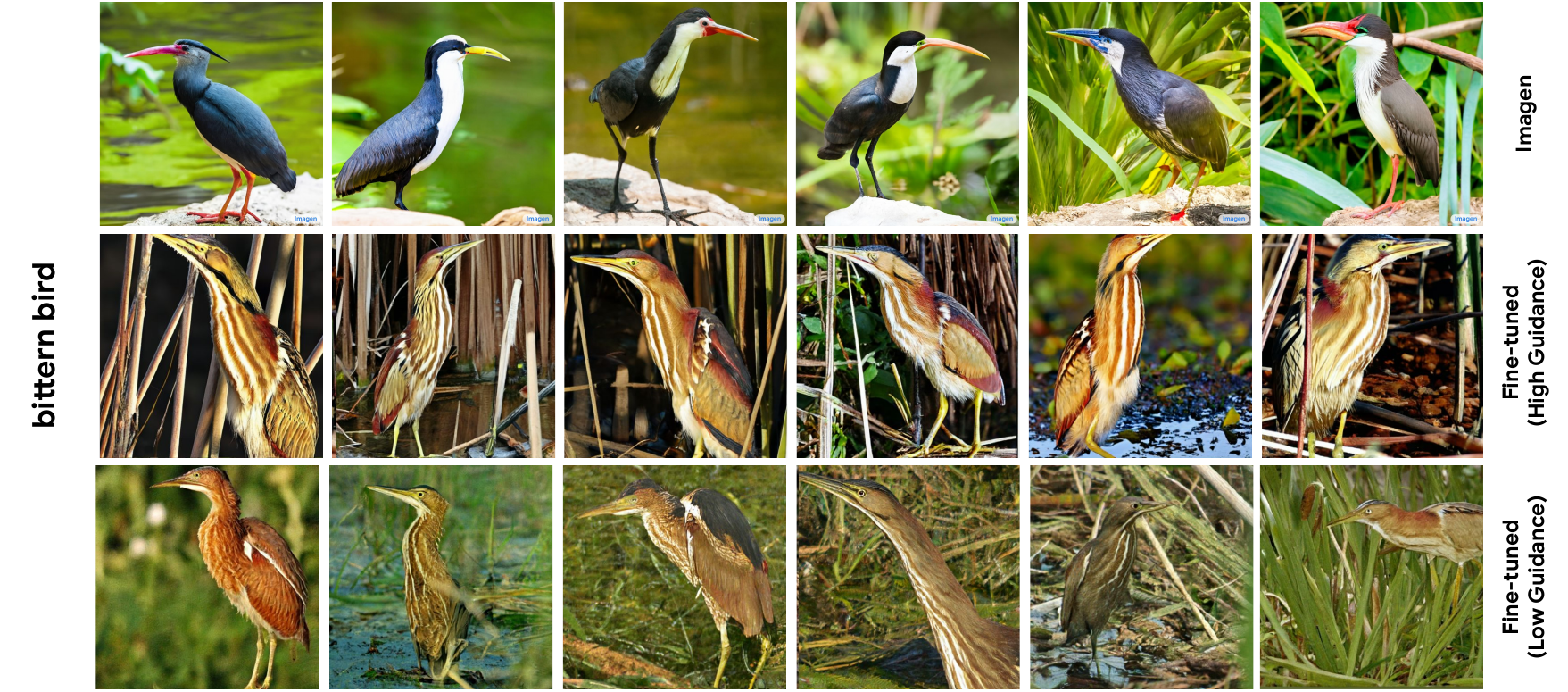} \\ \vspace{10pt}
\includegraphics[width=0.81\textwidth]{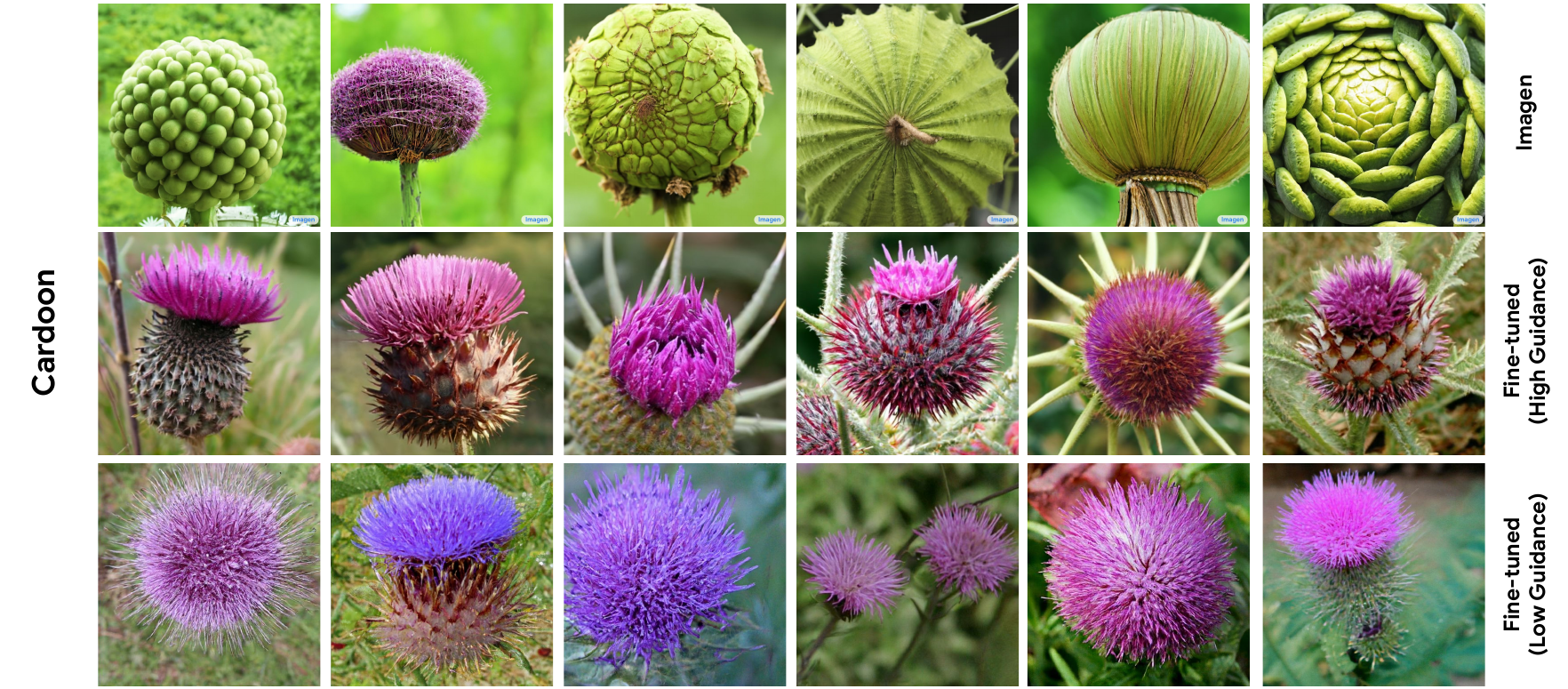} \\ \vspace{10pt} 
\includegraphics[width=0.81\textwidth]{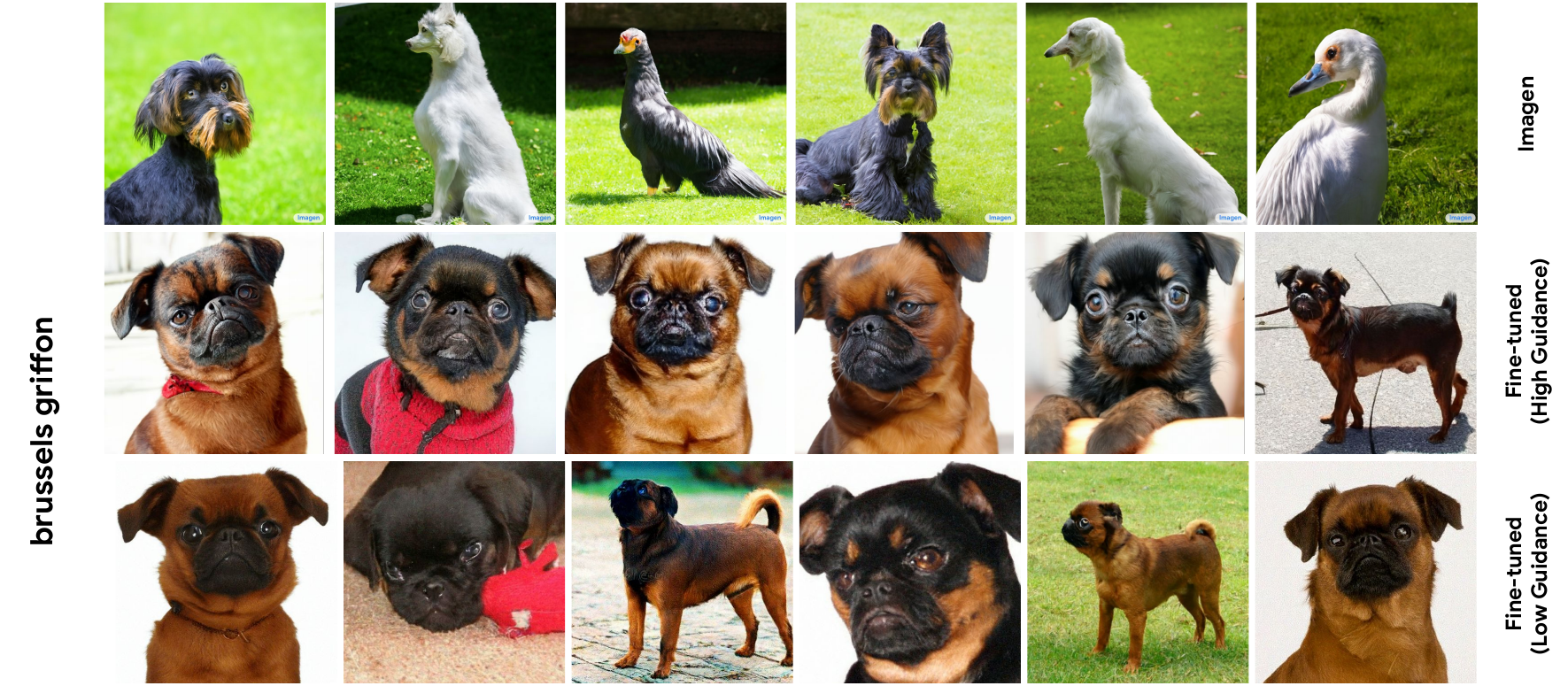} 
\end{center}
\vspace*{-0.25cm}
   \caption{Example $1024\!\times\! 1024$ images from vanilla Imagen (first row) vs.\ fine-tuned Imagen sampled with Imagen hyper-parameters (high guidance, second row) vs.\ fine-tuned Imagen sampled with our proposed hyper-parameter (low guidance, third row). Fine-tuning and careful choice of sampling parameters help to improve the alignment of images with class labels, and also improve sample diversity. Sampling with higher guidance weight can improve  photorealism, but lessens diversity.}
\label{fig:sample-image-com-part-1}
\end{figure}

\begin{figure}[h!]
\begin{center}
\includegraphics[width=0.81\textwidth]{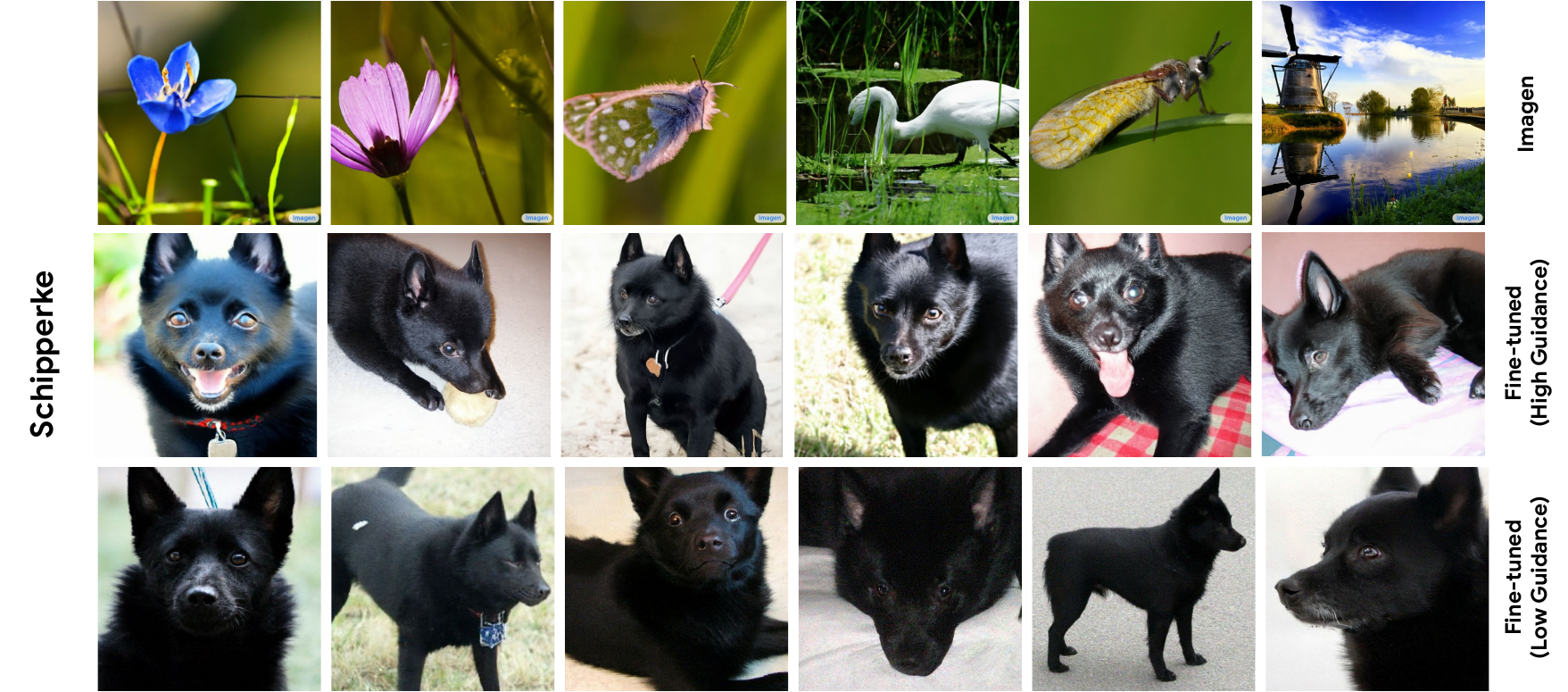} \\ \vspace{10pt}
\includegraphics[width=0.81\textwidth]{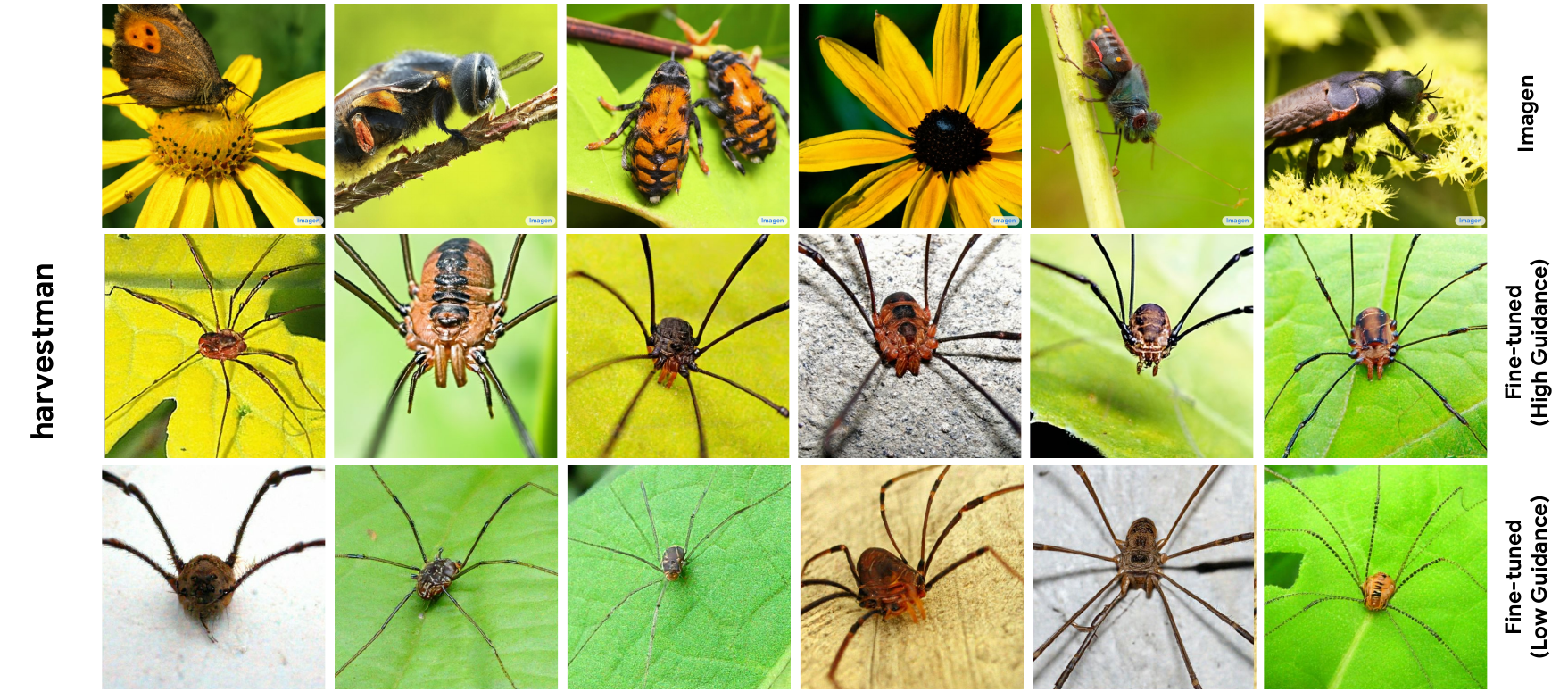} \\ \vspace{10pt}
\includegraphics[width=0.81\textwidth]{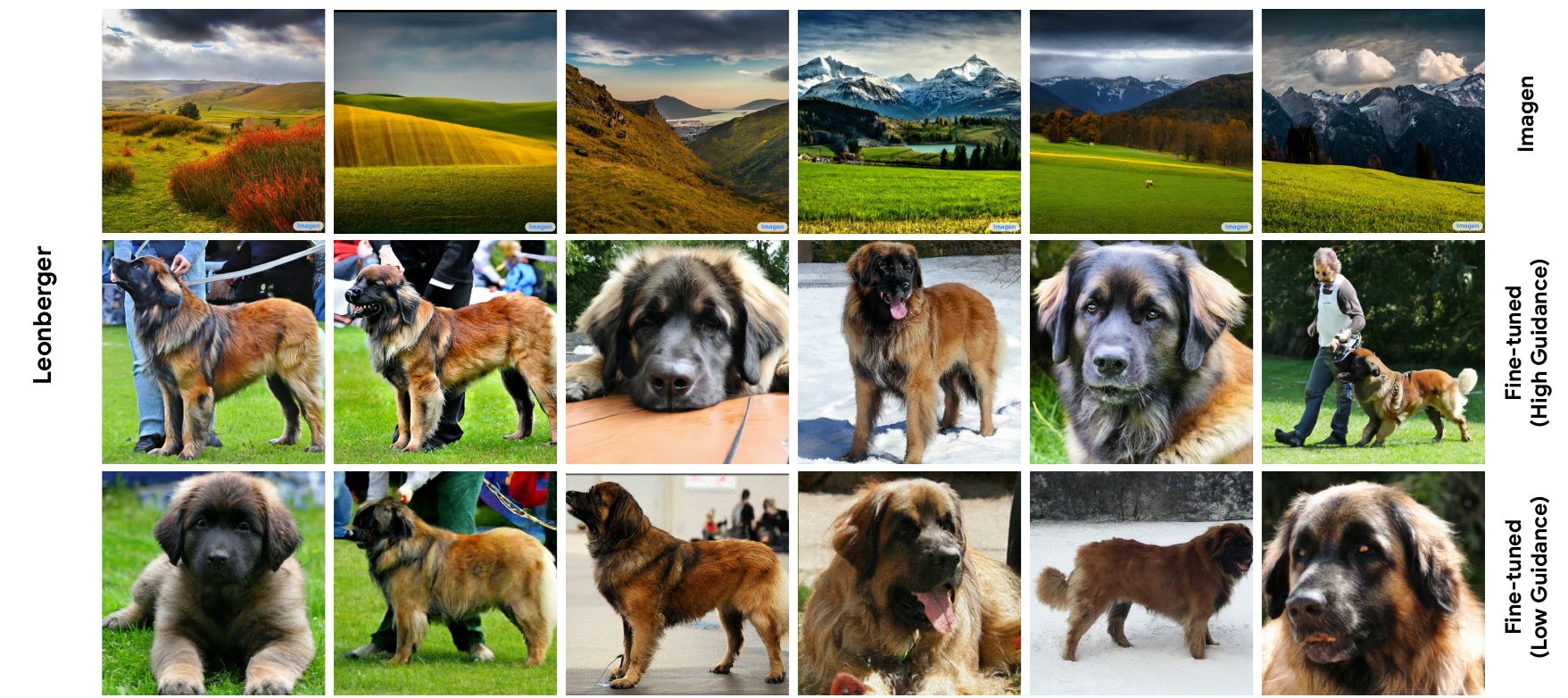} 
\end{center}
\vspace*{-0.25cm}
   \caption{Example $1024\!\times\! 1024$ images from vanilla Imagen (first row) vs.\ fine-tuned Imagen sampled with Imagen hyper-parameters (high guidance, second row) vs.\ fine-tuned Imagen sampled with our proposed hyper-parameter (low guidance, third row). Fine-tuning and careful choice of sampling parameters help to improve the alignment of images with class labels, and also improve sample diversity. Sampling with higher guidance weight can improve  photorealism, but lessens diversity.}
\label{fig:sample-image-com-part-2}
\end{figure}

\begin{figure}[h!]
\vspace*{1.0cm}
\begin{center}
\includegraphics[width=0.81\textwidth]{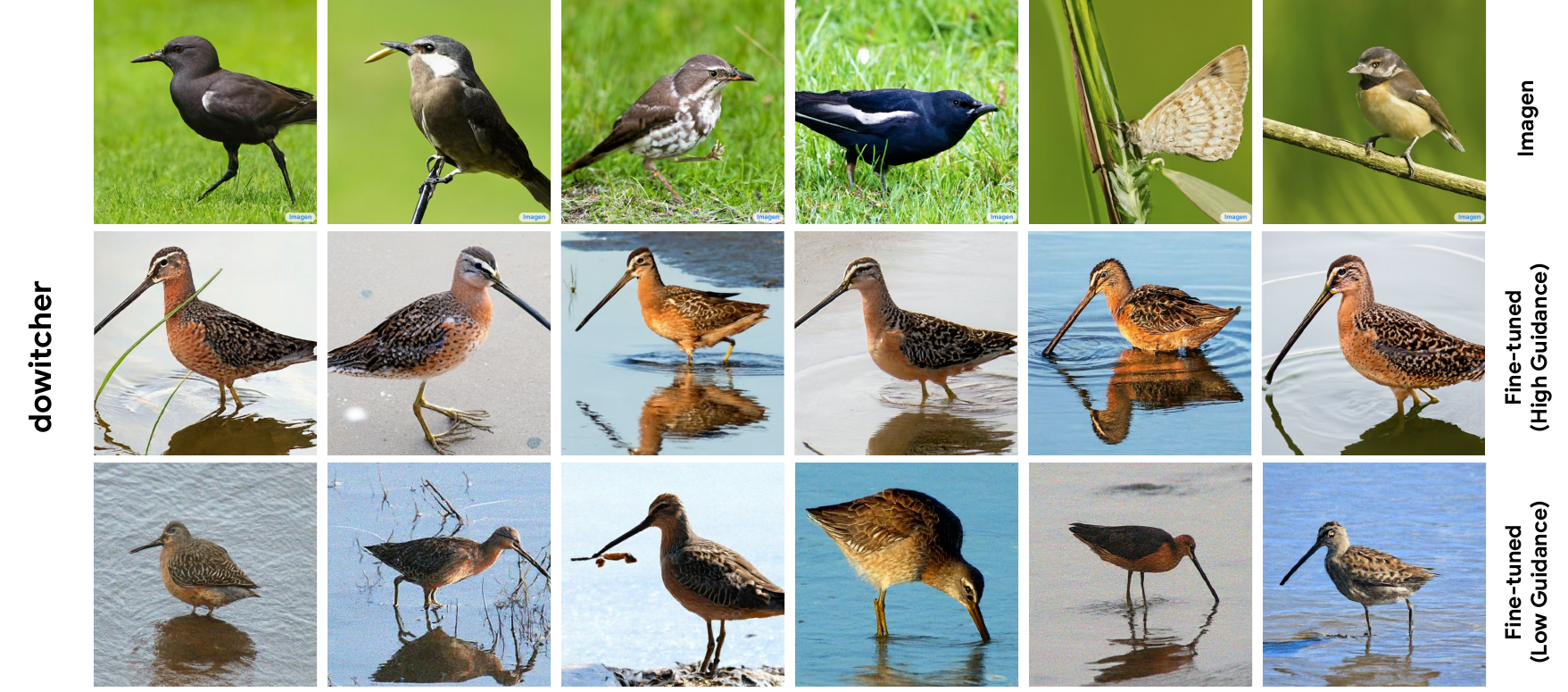} \\ \vspace{10pt}
\includegraphics[width=0.81\textwidth]{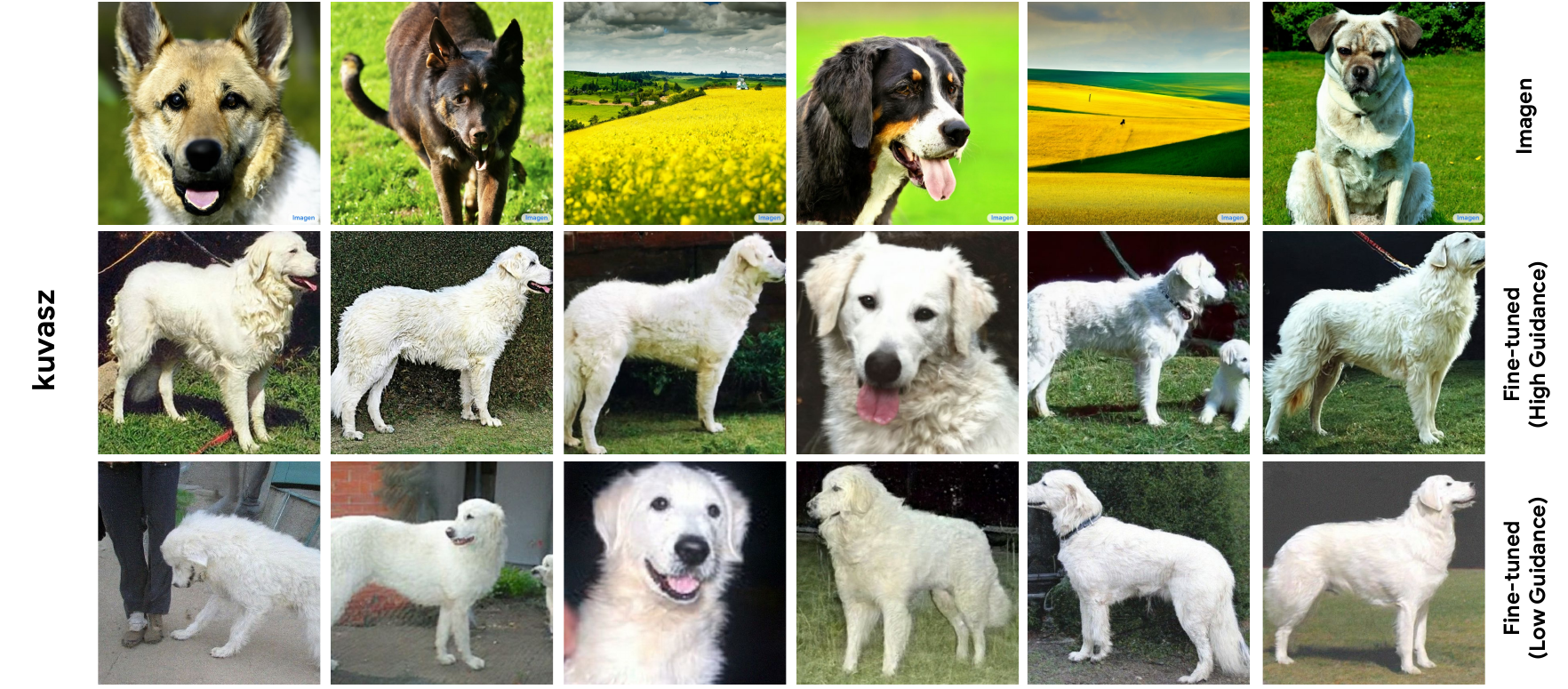} 
\end{center}
   \caption{Example $1024\!\times\! 1024$ images from vanilla Imagen (first row) vs.\ fine-tuned Imagen sampled with Imagen hyper-parameters (high guidance, second row) vs.\ fine-tuned Imagen sampled with our proposed hyper-parameter (low guidance, third row). Fine-tuning and careful choice of sampling parameters help to improve the alignment of images with class labels, and also improve sample diversity. Sampling with higher guidance weight can improve photorealism, but lessens diversity.}
   \vspace*{5.0cm}
\label{fig:sample-image-com-part-3}
\end{figure}

What follows are more samples to compare our fine-tuned model vs.\ the Imagen model are provided in Figure~\ref{fig:sample-image-com-part-1},~\ref{fig:sample-image-com-part-2}, and~\ref{fig:sample-image-com-part-3}. In this comparison we sample our fine-tuned model using two strategies. First, we sample using the proposed vanilla Imagen hyper-parameters which use a guidance weight of 10 for the sampling of the base $64\!\times\! 64$ model and subsequent super-resolution (SR) models are sampled with guidance weights of 20 and 8, respectively. This is called the high guidance strategy in these figures. Second, we use the proposed sampling hyper-parameters as explained in the paper which includes sampling the based model with a guidance weight of 1.25 and the subsequent SR models with a guidance weight of 1.0. This is called the low guidance weight strategy in these figures.

\vspace*{5cm}

\section{High Resolution Random Samples from the ImageNet Model}

\noindent

\begin{figure*}[h!]
\begin{center}
\includegraphics[width=0.98\textwidth]{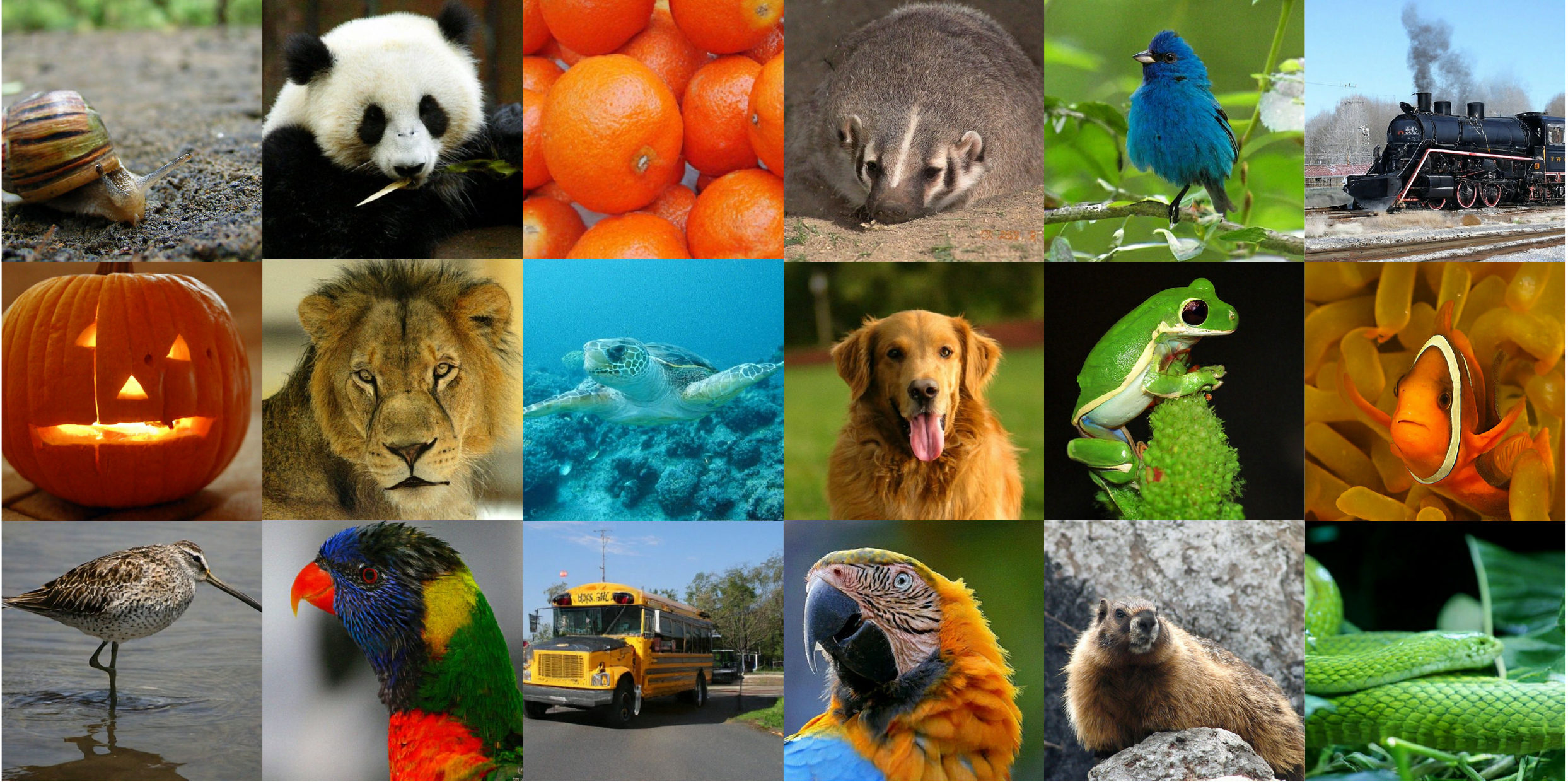}
\end{center}
\caption{ Random samples at 1024×1024 resolution generated by our fine-tuned model. The classes are snail (113), panda (388), orange (950), badger (362), indigo bunting (14), steam locomotive (820),  carved pumpkin (607), lion (291), loggerhead sea turtle (33),
golden retriever (207), tree frog (31),  clownfish (393),  dowitcher (142), lorikeet (90), school bus (779), macaw (88), marmot (336), green mamba (64).}
\label{fig:random-sample}
\end{figure*}

\newpage 

\section{Hyper-parameters and model selection for ImageNet classifiers.}
\label{app:hyperparameters}
This section details all the hyper-parameters used in training our ResNet-based model for CAS calculation, as well as the other ResNet-based, ResNet-RS-based, and Transformer-based models, used to report classifier accuracy in Table \ref{tab:data-augmentation}. Table \ref{tab:convnet-hyper} and Table \ref{tab:transformer-hyper} summarize the hyper-parameters used to train the ConvNet architectures and vision transformer architectures, respectively. 

For classification accuracy (CAS) calculation, as discussed before we follow the protocol  suggested in \cite{ravuri2019classification}. Our CAS ResNet-50 classifier is trained using a single crop. To train the classifier, we employ an SGD momentum optimizer and run it for 90 epochs. The learning rate is scheduled to linearly increase from 0.0 to 0.4 for the first five epochs and then decrease by a factor of 10 at epochs 30, 60, and 80. For other ResNet-based classifiers we employ more advanced mechanisms such as using a cosine schedule instead of step-wise learning rate decay, larger batch size, random augmentation, dropout, and label smoothing to reach competitive performance \cite{sun2017revisiting}.  It is also important to emphasize that ResNet-RS achieved higher performance than ResNet models through a combination of enhanced scaling strategies, improved training methodologies, and the implementation of techniques like the Squeeze-Excitation module \cite{bello2021revisiting}. We follow the training strategy and hyper-parameter suggested in \cite{bello2021revisiting} to train our ResNet-RS-based models. 

For vision transformer architectures we mainly follow the recipe provided in \cite{beyer2022better} to train a competitive ViT-S/16 model and \cite{touvron2021training} to train DeiT family models.  In all cases we re-implemented and train all of our models from scratch using real only, real + generated data, and generated only data until convergence.

\begin{table}[h]
\vspace*{0.4cm}
\centering
\small
\begin{tabular}{@{}l@{\hspace{.05cm}}|@{\hspace{.05cm}}ccccccc@{}}
\toprule
Model & ResNet-50 (CAS) & ResNet-50 & ResNet-101 & ResNet-152 & ResNet-RS-50 & ResNet-RS-101 & ResNet-RS-152 \\ \midrule
Epochs                \hspace{.05cm}& 90    & 130      & 200      & 200      & 350      & 350      & 350    \\ \midrule
Batch size            & 1024     & 4096     & 4096     & 4096     & 4096     & 4096     & 4096     \\
Optimizer             & Momentum & Momentum & Momentum & Momentum & Momentum & Momentum & Momentum \\
Learning rate         & 0.4      & 1.6      & 1.6      & 1.6      & 1.6      & 1.6      & 1.6      \\
Decay method          & Stepwise & Cosine   & Cosine   & Cosine   & Cosine   & Cosine   & Cosine   \\
Weight decay          & 1e-4     & 1e-4     & 1e-4     & 1e-4     & 4e-5     & 4e-5     & 4e-5     \\
Warmup epochs         & 5        & 5        & 5        & 5        & 5        & 5        & 5        \\
Label smoothing       & -        & 0.1      & 0.1      & 0.1      & 0.1      & 0.1      & 0.1      \\
Dropout rate          & -        & 0.25     & 0.25     & 0.25     & 0.25     & 0.25     & 0.25     \\
Rand Augment          & -        & 10       & 15       & 15       & 10       & 15       & 15       \\
\bottomrule
\end{tabular}
\caption{Hyper-parameters used to train ConvNet architectures including ResNet-50 (CAS) \cite{ravuri2019classification}, ResNet-50, ResNet-101, ResNet-152, ResNet-RS-50, ResNet-RS-101, and ResNet-RS-152 \cite{bello2021revisiting}.}
\label{tab:convnet-hyper}
\end{table}

\begin{table}[h]
\centering
\small
\begin{tabular}{@{}l|cccc}
\toprule
Model               & ViT-S/16 & DeiT-S & DeiT-B & DeiT-L \\ \midrule
Epochs              & 300      & 300    & 300    & 300    \\ \midrule
Batch size          & 1024     & 4096   & 4096   & 4096   \\
Optimizer           & AdamW    & AdamW  & AdamW  & AdamW  \\
Learning rate       & 0.001    & 0.004  & 0.004  & 0.004  \\
Learning rate decay & Cosine   & Cosine & Cosine & Cosine \\
Weight decay        & 0.0001   & -      & -      & -      \\ 
Warmup eepochs       & 10       & 5      & 5      & 5      \\ \midrule
Label dmoothing     & -        & 0.1    & 0.1    & 0.1    \\ \midrule
Rand Augment        & 10       & 9      & 9      & 9      \\
Mixup prob.         & 0.2      & 0.8    & 0.8    & 0.8    \\
Cutmix prob.        & -        & 1.0    & 1.0    & 1.0    \\
\bottomrule
\end{tabular}
\caption{Hyper-parameters used to train the vision transformer architectures, i.e., ViT-S/16~\cite{beyer2022better}, DeiT-S~\cite{touvron2021training}, DeiT-B~\cite{touvron2021training}, and DeiT-L~\cite{touvron2021training}.}
\label{tab:transformer-hyper}

\end{table}

\vfill

%% file: main.bbl
\begin{thebibliography}{10}\itemsep=-1pt

\bibitem{eDiff}
Yogesh Balaji, Seungjun Nah, Xun Huang, Arash Vahdat, Jiaming Song, Karsten
  Kreis, Miika Aittala, Timo Aila, Samuli Laine, Bryan Catanzaro, Tero Karras,
  and Ming-Yu Liu.
\newblock ediff-i: Text-to-image diffusion models with an ensemble of expert
  denoisers.
\newblock {\em preprint arxiv.2211.01324,}, 2022.

\bibitem{bansal2023leaving}
Hritik Bansal and Aditya Grover.
\newblock Leaving reality to imagination: Robust classification via generated
  datasets.
\newblock {\em arXiv preprint arXiv:2302.02503}, 2023.

\bibitem{baranchuk2021label}
Dmitry Baranchuk, Ivan Rubachev, Andrey Voynov, Valentin Khrulkov, and Artem
  Babenko.
\newblock Label-efficient semantic segmentation with diffusion models.
\newblock {\em arXiv preprint arXiv:2112.03126}, 2021.

\bibitem{bello2021revisiting}
Irwan Bello, William Fedus, Xianzhi Du, Ekin~Dogus Cubuk, Aravind Srinivas,
  Tsung-Yi Lin, Jonathon Shlens, and Barret Zoph.
\newblock Revisiting resnets: Improved training and scaling strategies.
\newblock {\em Advances in Neural Information Processing Systems},
  34:22614--22627, 2021.

\bibitem{beyer2022better}
Lucas Beyer, Xiaohua Zhai, and Alexander Kolesnikov.
\newblock Better plain vit baselines for imagenet-1k.
\newblock {\em arXiv preprint arXiv:2205.01580}, 2022.

\bibitem{brock2019large}
Andrew Brock, Jeff Donahue, and Karen Simonyan.
\newblock Large scale {GAN} training for high fidelity natural image synthesis.
\newblock {\em International Conference on Learning Representations}, 2019.

\bibitem{bucilua2006model}
Cristian Buciluǎ, Rich Caruana, and Alexandru Niculescu-Mizil.
\newblock Model compression.
\newblock In {\em Proceedings of the 12th ACM SIGKDD international conference
  on Knowledge discovery and data mining}, pages 535--541, 2006.

\bibitem{chen2020wavegrad}
Nanxin Chen, Yu Zhang, Heiga Zen, Ron~J Weiss, Mohammad Norouzi, and William
  Chan.
\newblock Wavegrad: Estimating gradients for waveform generation.
\newblock {\em arXiv preprint arXiv:2009.00713}, 2020.

\bibitem{chen2023importance}
Ting Chen.
\newblock On the importance of noise scheduling for diffusion models.
\newblock {\em arXiv preprint arXiv:2301.10972}, 2023.

\bibitem{chen2019learning}
Yuhua Chen, Wen Li, Xiaoran Chen, and Luc~Van Gool.
\newblock Learning semantic segmentation from synthetic data: A geometrically
  guided input-output adaptation approach.
\newblock In {\em Proceedings of the IEEE/CVF conference on computer vision and
  pattern recognition}, pages 1841--1850, 2019.

\bibitem{de2021next}
Celso~M de Melo, Antonio Torralba, Leonidas Guibas, James DiCarlo, Rama
  Chellappa, and Jessica Hodgins.
\newblock Next-generation deep learning based on simulators and synthetic data.
\newblock {\em Trends in cognitive sciences}, 2021.

\bibitem{dhariwal2021diffusion}
Prafulla Dhariwal and Alexander Nichol.
\newblock Diffusion models beat {GANs} on image synthesis.
\newblock {\em Advances in Neural Information Processing Systems},
  34:8780--8794, 2021.

\bibitem{dosovitskiy2020image}
Alexey Dosovitskiy, Lucas Beyer, Alexander Kolesnikov, Dirk Weissenborn,
  Xiaohua Zhai, Thomas Unterthiner, Mostafa Dehghani, Matthias Minderer, Georg
  Heigold, Sylvain Gelly, et~al.
\newblock An image is worth 16x16 words: Transformers for image recognition at
  scale.
\newblock {\em arXiv preprint arXiv:2010.11929}, 2020.

\bibitem{dosovitskiy2015flownet}
Alexey Dosovitskiy, Philipp Fischer, Eddy Ilg, Philip Hausser, Caner Hazirbas,
  Vladimir Golkov, Patrick Van Der~Smagt, Daniel Cremers, and Thomas Brox.
\newblock Flownet: Learning optical flow with convolutional networks.
\newblock In {\em Proceedings of the IEEE international conference on computer
  vision}, pages 2758--2766, 2015.

\bibitem{dosovitskiy2017carla}
Alexey Dosovitskiy, German Ros, Felipe Codevilla, Antonio Lopez, and Vladlen
  Koltun.
\newblock Carla: An open urban driving simulator.
\newblock In {\em Conference on robot learning}, pages 1--16. PMLR, 2017.

\bibitem{gan2021threedworld}
C Gan, J Schwartz, S Alter, M Schrimpf, J Traer, J De~Freitas, J Kubilius, A
  Bhandwaldar, N Haber, M Sano, et~al.
\newblock Threedworld: A platform for interactive multi-modal physical
  simulation.
\newblock {\em Advances in Neural Information Processing Systems (NeurIPS)},
  2021.

\bibitem{gowal2021improving}
Sven Gowal, Sylvestre-Alvise Rebuffi, Olivia Wiles, Florian Stimberg,
  Dan~Andrei Calian, and Timothy~A Mann.
\newblock Improving robustness using generated data.
\newblock {\em Advances in Neural Information Processing Systems},
  34:4218--4233, 2021.

\bibitem{greff2021kubric}
Klaus Greff, Francois Belletti, Lucas Beyer, Carl Doersch, Yilun Du, Daniel
  Duckworth, David~J Fleet, Dan Gnanapragasam, Florian Golemo, Charles
  Herrmann, Thomas Kipf, Abhijit Kundu, Dmitry Lagun, Issam Laradji,
  Hsueh-Ti~(Derek) Liu, Henning Meyer, Yishu Miao, Derek Nowrouzezahrai, Cengiz
  Oztireli, Etienne Pot, Noha Radwan, Daniel Rebain, Sara Sabour, Mehdi S.~M.
  Sajjadi, Matan Sela, Vincent Sitzmann, Austin Stone, Deqing Sun, Suhani Vora,
  Ziyu Wang, Tianhao Wu, Kwang~Moo Yi, Fangcheng Zhong, and Andrea
  Tagliasacchi.
\newblock Kubric: A scalable dataset generator.
\newblock 2022.

\bibitem{guo2022learning}
Xi Guo, Wei Wu, Dongliang Wang, Jing Su, Haisheng Su, Weihao Gan, Jian Huang,
  and Qin Yang.
\newblock Learning video representations of human motion from synthetic data.
\newblock In {\em Proceedings of the IEEE/CVF Conference on Computer Vision and
  Pattern Recognition}, pages 20197--20207, 2022.

\bibitem{he2016deep}
Kaiming He, Xiangyu Zhang, Shaoqing Ren, and Jian Sun.
\newblock Deep residual learning for image recognition.
\newblock In {\em Proceedings of the IEEE conference on computer vision and
  pattern recognition}, pages 770--778, 2016.

\bibitem{he2022synthetic}
Ruifei He, Shuyang Sun, Xin Yu, Chuhui Xue, Wenqing Zhang, Philip Torr, Song
  Bai, and Xiaojuan Qi.
\newblock Is synthetic data from generative models ready for image recognition?
\newblock {\em arXiv preprint arXiv:2210.07574}, 2022.

\bibitem{heusel2017gans}
Martin Heusel, Hubert Ramsauer, Thomas Unterthiner, Bernhard Nessler, and Sepp
  Hochreiter.
\newblock Gans trained by a two time-scale update rule converge to a local nash
  equilibrium.
\newblock {\em Advances in neural information processing systems}, 30, 2017.

\bibitem{hinton2015distilling}
Geoffrey Hinton, Oriol Vinyals, and Jeff Dean.
\newblock Distilling the knowledge in a neural network.
\newblock {\em arXiv preprint arXiv:1503.02531}, 2015.

\bibitem{ho2022imagen}
Jonathan Ho, William Chan, Chitwan Saharia, Jay Whang, Ruiqi Gao, Alexey
  Gritsenko, Diederik~P Kingma, Ben Poole, Mohammad Norouzi, David~J Fleet,
  et~al.
\newblock Imagen video: High definition video generation with diffusion models.
\newblock {\em arXiv preprint arXiv:2210.02303}, 2022.

\bibitem{ho2020denoising}
Jonathan Ho, Ajay Jain, and Pieter Abbeel.
\newblock Denoising diffusion probabilistic models.
\newblock {\em Advances in Neural Information Processing Systems},
  33:6840--6851, 2020.

\bibitem{ho2022cascaded}
Jonathan Ho, Chitwan Saharia, William Chan, David~J Fleet, Mohammad Norouzi,
  and Tim Salimans.
\newblock Cascaded diffusion models for high fidelity image generation.
\newblock {\em J. Mach. Learn. Res.}, 23(47):1--33, 2022.

\bibitem{ho2022classifier}
Jonathan Ho and Tim Salimans.
\newblock Classifier-free diffusion guidance.
\newblock {\em arXiv preprint arXiv:2207.12598}, 2022.

\bibitem{hoogeboom2023simple}
Emiel Hoogeboom, Jonathan Heek, and Tim Salimans.
\newblock simple diffusion: End-to-end diffusion for high resolution images.
\newblock {\em arXiv preprint arXiv:2301.11093}, 2023.

\bibitem{izadi2011kinectfusion}
Shahram Izadi, David Kim, Otmar Hilliges, David Molyneaux, Richard Newcombe,
  Pushmeet Kohli, Jamie Shotton, Steve Hodges, Dustin Freeman, Andrew Davison,
  et~al.
\newblock Kinectfusion: real-time 3d reconstruction and interaction using a
  moving depth camera.
\newblock In {\em Proceedings of the 24th annual ACM symposium on User
  interface software and technology}, pages 559--568, 2011.

\bibitem{jabri2022scalable}
Allan Jabri, David Fleet, and Ting Chen.
\newblock Scalable adaptive computation for iterative generation.
\newblock {\em arXiv preprint arXiv:2212.11972}, 2022.

\bibitem{KarrasNeurIPS2022}
Tero Karras, Miika Aittala, Timo Aila, and Samuli Laine.
\newblock Elucidating the design space of diffusion-based generative models.
\newblock {\em NeurIPS}, 2022.

\bibitem{kim2022transferable}
Yo-whan Kim.
\newblock {\em How Transferable are Video Representations Based on Synthetic
  Data?}
\newblock PhD thesis, Massachusetts Institute of Technology, 2022.

\bibitem{adam}
Diederik~P. Kingma and Jimmy Ba.
\newblock Adam: A method for stochastic optimization.
\newblock {\em arXiv preprint arxiv:1412.6980}, 2014.

\bibitem{kolesnikov2020big}
Alexander Kolesnikov, Lucas Beyer, Xiaohua Zhai, Joan Puigcerver, Jessica Yung,
  Sylvain Gelly, and Neil Houlsby.
\newblock Big transfer (bit): General visual representation learning.
\newblock In {\em Computer Vision--ECCV 2020: 16th European Conference,
  Glasgow, UK, August 23--28, 2020, Proceedings, Part V 16}, pages 491--507.
  Springer, 2020.

\bibitem{kong2020diffwave}
Zhifeng Kong, Wei Ping, Jiaji Huang, Kexin Zhao, and Bryan Catanzaro.
\newblock Diffwave: A versatile diffusion model for audio synthesis.
\newblock {\em arXiv preprint arXiv:2009.09761}, 2020.

\bibitem{li2022bigdatasetgan}
Daiqing Li, Huan Ling, Seung~Wook Kim, Karsten Kreis, Sanja Fidler, and Antonio
  Torralba.
\newblock Bigdatasetgan: Synthesizing imagenet with pixel-wise annotations.
\newblock In {\em Proceedings of the IEEE/CVF Conference on Computer Vision and
  Pattern Recognition}, pages 21330--21340, 2022.

\bibitem{li2021semantic}
Daiqing Li, Junlin Yang, Karsten Kreis, Antonio Torralba, and Sanja Fidler.
\newblock Semantic segmentation with generative models: Semi-supervised
  learning and strong out-of-domain generalization.
\newblock In {\em Proceedings of the IEEE/CVF Conference on Computer Vision and
  Pattern Recognition}, pages 8300--8311, 2021.

\bibitem{ma2022pretrained}
Jianxin Ma, Shuai Bai, and Chang Zhou.
\newblock Pretrained diffusion models for unified human motion synthesis.
\newblock {\em arXiv preprint arXiv:2212.02837}, 2022.

\bibitem{mahajan2018exploring}
Dhruv Mahajan, Ross Girshick, Vignesh Ramanathan, Kaiming He, Manohar Paluri,
  Yixuan Li, Ashwin Bharambe, and Laurens Van Der~Maaten.
\newblock Exploring the limits of weakly supervised pretraining.
\newblock In {\em Proceedings of the European conference on computer vision
  (ECCV)}, pages 181--196, 2018.

\bibitem{nichol2021glide}
Alex Nichol, Prafulla Dhariwal, Aditya Ramesh, Pranav Shyam, Pamela Mishkin,
  Bob McGrew, Ilya Sutskever, and Mark Chen.
\newblock Glide: Towards photorealistic image generation and editing with
  text-guided diffusion models.
\newblock {\em arXiv preprint arXiv:2112.10741}, 2021.

\bibitem{nichol2021improved}
Alexander~Quinn Nichol and Prafulla Dhariwal.
\newblock Improved denoising diffusion probabilistic models.
\newblock In {\em International Conference on Machine Learning}, pages
  8162--8171. PMLR, 2021.

\bibitem{peebles2022gan}
William Peebles, Jun-Yan Zhu, Richard Zhang, Antonio Torralba, Alexei~A Efros,
  and Eli Shechtman.
\newblock Gan-supervised dense visual alignment.
\newblock In {\em Proceedings of the IEEE/CVF Conference on Computer Vision and
  Pattern Recognition}, pages 13470--13481, 2022.

\bibitem{radford2021learning}
Alec Radford, Jong~Wook Kim, Chris Hallacy, Aditya Ramesh, Gabriel Goh,
  Sandhini Agarwal, Girish Sastry, Amanda Askell, Pamela Mishkin, Jack Clark,
  et~al.
\newblock Learning transferable visual models from natural language
  supervision.
\newblock In {\em International conference on machine learning}, pages
  8748--8763. PMLR, 2021.

\bibitem{ramesh2022hierarchical}
Aditya Ramesh, Prafulla Dhariwal, Alex Nichol, Casey Chu, and Mark Chen.
\newblock Hierarchical text-conditional image generation with clip latents.
\newblock {\em arXiv preprint arXiv:2204.06125}, 2022.

\bibitem{ravuri2019classification}
Suman Ravuri and Oriol Vinyals.
\newblock Classification accuracy score for conditional generative models.
\newblock {\em Advances in neural information processing systems}, 32, 2019.

\bibitem{razavi2019generating}
Ali Razavi, Aaron Van~den Oord, and Oriol Vinyals.
\newblock Generating diverse high-fidelity images with vq-vae-2.
\newblock {\em Advances in neural information processing systems}, 32, 2019.

\bibitem{rombach2022high}
Robin Rombach, Andreas Blattmann, Dominik Lorenz, Patrick Esser, and Bj{\"o}rn
  Ommer.
\newblock High-resolution image synthesis with latent diffusion models.
\newblock In {\em Proceedings of the IEEE/CVF Conference on Computer Vision and
  Pattern Recognition}, pages 10684--10695, 2022.

\bibitem{russakovsky2015imagenet}
Olga Russakovsky, Jia Deng, Hao Su, Jonathan Krause, Sanjeev Satheesh, Sean Ma,
  Zhiheng Huang, Andrej Karpathy, Aditya Khosla, Michael Bernstein, et~al.
\newblock Imagenet large scale visual recognition challenge.
\newblock {\em International journal of computer vision}, 115:211--252, 2015.

\bibitem{saharia2022palette}
Chitwan Saharia, William Chan, Huiwen Chang, Chris Lee, Jonathan Ho, Tim
  Salimans, David Fleet, and Mohammad Norouzi.
\newblock Palette: Image-to-image diffusion models.
\newblock In {\em ACM SIGGRAPH 2022 Conference Proceedings}, pages 1--10, 2022.

\bibitem{saharia2022photorealistic}
Chitwan Saharia, William Chan, Saurabh Saxena, Lala Li, Jay Whang, Emily
  Denton, Seyed Kamyar~Seyed Ghasemipour, Burcu~Karagol Ayan, S~Sara Mahdavi,
  Rapha~Gontijo Lopes, et~al.
\newblock Photorealistic text-to-image diffusion models with deep language
  understanding.
\newblock {\em arXiv preprint arXiv:2205.11487}, 2022.

\bibitem{saharia2022image}
Chitwan Saharia, Jonathan Ho, William Chan, Tim Salimans, David~J Fleet, and
  Mohammad Norouzi.
\newblock Image super-resolution via iterative refinement.
\newblock {\em IEEE Transactions on Pattern Analysis and Machine Intelligence},
  2022.

\bibitem{salimans2016improved}
Tim Salimans, Ian Goodfellow, Wojciech Zaremba, Vicki Cheung, Alec Radford, and
  Xi Chen.
\newblock Improved techniques for training gans.
\newblock {\em Advances in neural information processing systems}, 29, 2016.

\bibitem{salimans2022progressive}
Tim Salimans and Jonathan Ho.
\newblock Progressive distillation for fast sampling of diffusion models.
\newblock {\em arXiv preprint arXiv:2202.00512}, 2022.

\bibitem{sankaranarayanan2018learning}
Swami Sankaranarayanan, Yogesh Balaji, Arpit Jain, Ser~Nam Lim, and Rama
  Chellappa.
\newblock Learning from synthetic data: Addressing domain shift for semantic
  segmentation.
\newblock In {\em Proceedings of the IEEE conference on computer vision and
  pattern recognition}, pages 3752--3761, 2018.

\bibitem{santurkar2018classification}
Shibani Santurkar, Ludwig Schmidt, and Aleksander Madry.
\newblock A classification-based study of covariate shift in gan distributions.
\newblock In {\em International Conference on Machine Learning}, pages
  4480--4489. PMLR, 2018.

\bibitem{sariyildiz2022fake}
Mert~Bulent Sariyildiz, Karteek Alahari, Diane Larlus, and Yannis Kalantidis.
\newblock Fake it till you make it: Learning (s) from a synthetic imagenet
  clone.
\newblock {\em arXiv preprint arXiv:2212.08420}, 2022.

\bibitem{adafactor}
Noam Shazeer and Mitchell Stern.
\newblock Adafactor: Adaptive learning rates with sublinear memory cost.
\newblock {\em arXiv preprint, arxiv:1804.04235}, 2018.

\bibitem{singer2022make}
Uriel Singer, Adam Polyak, Thomas Hayes, Xi Yin, Jie An, Songyang Zhang, Qiyuan
  Hu, Harry Yang, Oron Ashual, Oran Gafni, et~al.
\newblock Make-a-video: Text-to-video generation without text-video data.
\newblock {\em arXiv preprint arXiv:2209.14792}, 2022.

\bibitem{sohl2015deep}
Jascha Sohl-Dickstein, Eric Weiss, Niru Maheswaranathan, and Surya Ganguli.
\newblock Deep unsupervised learning using nonequilibrium thermodynamics.
\newblock In {\em International Conference on Machine Learning}, pages
  2256--2265. PMLR, 2015.

\bibitem{song2020denoising}
Jiaming Song, Chenlin Meng, and Stefano Ermon.
\newblock Denoising diffusion implicit models.
\newblock {\em International Confernece on Learning Representations}, 2021.

\bibitem{song2020score}
Yang Song, Jascha Sohl-Dickstein, Diederik~P Kingma, Abhishek Kumar, Stefano
  Ermon, and Ben Poole.
\newblock Score-based generative modeling through stochastic differential
  equations.
\newblock {\em arXiv preprint arXiv:2011.13456}, 2020.

\bibitem{sun2017revisiting}
Chen Sun, Abhinav Shrivastava, Saurabh Singh, and Abhinav Gupta.
\newblock Revisiting unreasonable effectiveness of data in deep learning era.
\newblock In {\em Proceedings of the IEEE international conference on computer
  vision}, pages 843--852, 2017.

\bibitem{sun2021autoflow}
Deqing Sun, Daniel Vlasic, Charles Herrmann, Varun Jampani, Michael Krainin,
  Huiwen Chang, Ramin Zabih, William~T Freeman, and Ce Liu.
\newblock Autoflow: Learning a better training set for optical flow.
\newblock In {\em Proceedings of the IEEE/CVF Conference on Computer Vision and
  Pattern Recognition}, pages 10093--10102, 2021.

\bibitem{touvron2021training}
Hugo Touvron, Matthieu Cord, Matthijs Douze, Francisco Massa, Alexandre
  Sablayrolles, and Herv{\'e} J{\'e}gou.
\newblock Training data-efficient image transformers \& distillation through
  attention.
\newblock In {\em International conference on machine learning}, pages
  10347--10357. PMLR, 2021.

\bibitem{trabucco2023effective}
Brandon Trabucco, Kyle Doherty, Max Gurinas, and Ruslan Salakhutdinov.
\newblock Effective data augmentation with diffusion models.
\newblock {\em arXiv preprint arXiv:2302.07944}, 2023.

\bibitem{tritrong2021repurposing}
Nontawat Tritrong, Pitchaporn Rewatbowornwong, and Supasorn Suwajanakorn.
\newblock Repurposing gans for one-shot semantic part segmentation.
\newblock In {\em Proceedings of the IEEE/CVF conference on computer vision and
  pattern recognition}, pages 4475--4485, 2021.

\bibitem{varol2017learning}
Gul Varol, Javier Romero, Xavier Martin, Naureen Mahmood, Michael~J Black, Ivan
  Laptev, and Cordelia Schmid.
\newblock Learning from synthetic humans.
\newblock In {\em Proceedings of the IEEE conference on computer vision and
  pattern recognition}, pages 109--117, 2017.

\bibitem{villegas2022phenaki}
Ruben Villegas, Mohammad Babaeizadeh, Pieter-Jan Kindermans, Hernan Moraldo,
  Han Zhang, Mohammad~Taghi Saffar, Santiago Castro, Julius Kunze, and Dumitru
  Erhan.
\newblock Phenaki: Variable length video generation from open domain textual
  description.
\newblock {\em arXiv preprint arXiv:2210.02399}, 2022.

\bibitem{wang2022imagen}
Su Wang, Chitwan Saharia, Ceslee Montgomery, Jordi Pont-Tuset, Shai Noy,
  Stefano Pellegrini, Yasumasa Onoe, Sarah Laszlo, David~J Fleet, Radu Soricut,
  et~al.
\newblock Imagen editor and editbench: Advancing and evaluating text-guided
  image inpainting.
\newblock {\em arXiv preprint arXiv:2212.06909}, 2022.

\bibitem{xu2021generative}
Yinghao Xu, Yujun Shen, Jiapeng Zhu, Ceyuan Yang, and Bolei Zhou.
\newblock Generative hierarchical features from synthesizing images.
\newblock In {\em Proceedings of the IEEE/CVF Conference on Computer Vision and
  Pattern Recognition}, pages 4432--4442, 2021.

\bibitem{yang2017lrgan}
Jianwei Yang, Anitha Kannan, Dhruv Batra, and Devi Parikh.
\newblock {LR}-{GAN}: Layered recursive generative adversarial networks for
  image generation.
\newblock In {\em International Conference on Learning Representations}, 2017.

\bibitem{zhai2022scaling}
Xiaohua Zhai, Alexander Kolesnikov, Neil Houlsby, and Lucas Beyer.
\newblock Scaling vision transformers.
\newblock In {\em Proceedings of the IEEE/CVF Conference on Computer Vision and
  Pattern Recognition}, pages 12104--12113, 2022.

\bibitem{zheng2020structured3d}
Jia Zheng, Junfei Zhang, Jing Li, Rui Tang, Shenghua Gao, and Zihan Zhou.
\newblock Structured3d: A large photo-realistic dataset for structured 3d
  modeling.
\newblock In {\em Computer Vision--ECCV 2020: 16th European Conference,
  Glasgow, UK, August 23--28, 2020, Proceedings, Part IX 16}, pages 519--535.
  Springer, 2020.

\end{thebibliography}
